\documentclass[times, review, 10pt]{elsarticle}

\usepackage{tabularx}   
\usepackage{makecell}   
\usepackage[utf8]{inputenc} 
\usepackage{xcolor}         
\usepackage{xfp}            
\usepackage{etoolbox}      
\usepackage{tcolorbox}

\usepackage{amssymb}
\usepackage{amsmath}
\usepackage{amsthm} 

\usepackage{booktabs}
\usepackage{siunitx}
\usepackage[hidelinks, colorlinks=true]{hyperref} 
\usepackage{caption}    
\usepackage{subcaption} 
\usepackage{seqsplit}

\DeclareCaptionLabelFormat{myformat}{\textbf{Fig. #2}}
\newcommand{\eqautoref}[1]{\hyperref[#1]{Eq.(\ref*{#1})}}

\definecolor{myred}{rgb}{0.8, 0, 0}
\definecolor{Gray}{gray}{0.95}
\definecolor{darkgray}{gray}{0.6} 

\newcommand{\showgain}[2]{
    \textcolor{myred}{
        \footnotesize
        \ensuremath{\uparrow}\fpeval{round(((#1/#2)-1)*100, 1)}\%
    }%
}
\robustify{\showgain}
\newcommand{\showPerfChange}[2]{%
  \num[round-mode=places, round-precision=2]{#1}%
  \ifdim\fpeval{#1}pt > \fpeval{#2}pt\relax
    \textcolor{myred}{\tiny\ (\ensuremath{\uparrow}\num[round-mode=places, round-precision=2]{\fpeval{#1 - #2}})}%
  \fi
}

\journal{Pattern Recognition}

\begin{document}

\begin{frontmatter}



\title{High-Quality Proposal Encoding and Cascade Denoising for Imaginary Supervised Object Detection}

\author[hitsz]{Zhiyuan Chen}
\ead{xeesoxeechen@gmail.com}
\author[hitsz]{Yuelin Guo}
\ead{gyl2565309278@gmail.com}
\author[hit2]{Zitong Huang}
\ead{zitonghuang@outlook.com}
\author[mu]{Haoyu He}
\ead{Charles.haoyu.he@gmail.com}
\author[pcl]{Renhao Lu}
\ead{lurh100@pcl.ac.cn}
\author[hitsz,pcl,hit1]{Weizhe Zhang\corref{cor1}}
\ead{wzzhang@hit.edu.cn}
\cortext[cor1]{Corresponding author.}
\affiliation[hitsz]{
    organization={Institute of Cyberspace Security, Harbin Institute of Technology, Shenzhen},
    addressline={University Town of Shenzhen, Nanshan District},
    city={Shenzhen},
    postcode={518055},
    state={Guangdong},
    country={China},
}
\affiliation[pcl]{
    organization={Department of New Networks, Peng Cheng Laboratory},
    addressline={Xili Lake International Science and Education City, Nanshan District},
    city={Shenzhen},
    postcode={518066},
    state={Guangdong},
    country={China},
}
\affiliation[hit1]{
    organization={School of Cyberspace Science, Harbin Institute of Technology},
    addressline={No. 92 West Dazhi Street, Nangang District},
    city={Harbin},
    postcode={150001},
    state={Heilongjiang},
    country={China},
}
\affiliation[hit2]{
    organization={Center on Machine Learning Research, Harbin Institute of Technology},
    addressline={No. 92 West Dazhi Street, Nangang District},
    city={Harbin},
    postcode={150001},
    state={Heilongjiang},
    country={China},
}
\affiliation[mu]{
    organization={Faculty of Information Technology, Monash University},
    addressline={Wellington Road},
    city={Clayton},
    postcode={3800},
    state={Victoria},
    country={Australia},
}

\begin{abstract}
Object detection models demand large-scale annotated datasets, which are costly and labor-intensive to create. This motivated Imaginary Supervised Object Detection (ISOD), where models train on synthetic images and test on real images. However, existing methods face three limitations: (1) synthetic datasets suffer from simplistic prompts, poor image quality, and weak supervision; (2) DETR-based detectors, due to their random query initialization, struggle with slow convergence and overfitting to synthetic patterns, hindering real-world generalization; (3) uniform denoising pressure promotes model overfitting to pseudo-label noise. We propose Cascade HQP-DETR to address these limitations. First, we introduce a high-quality data pipeline using LLaMA-3, Flux, and Grounding DINO to generate the FluxVOC and FluxCOCO datasets, advancing ISOD from weak to full supervision. Second, our High-Quality Proposal guided query encoding initializes object queries with image-specific priors from SAM-generated proposals and RoI-pooled features, accelerating convergence while steering the model to learn transferable features instead of overfitting to synthetic patterns. Third, our cascade denoising algorithm dynamically adjusts training weights through progressively increasing IoU thresholds across decoder layers, guiding the model to learn robust boundaries from reliable visual cues rather than overfitting to noisy labels. Trained for just 12 epochs solely on FluxVOC, Cascade HQP-DETR achieves a SOTA 61.04\% mAP@0.5 on PASCAL VOC 2007, outperforming strong baselines, with its competitive real-data performance confirming the architecture's universal applicability.
\end{abstract}
\end{frontmatter}
\section{Introduction}
\label{Introduction}
Object detection is a fundamental computer vision task that identifies and localizes objects, with applications across multiple domains \cite{COD-SAM,yolopx}. Driven by deep learning advances \cite{resnet,vit}, detection models \cite{cascadercnn,fasterrcnn,yolo9} have progressed substantially. However, these models require large-scale manually annotated datasets \cite{coco,pascalvoc} which are expensive and labor-intensive. To address this, Imaginary Supervised Object Detection (ISOD) \cite{imaginarynet} offers an alternative where models train on synthetic images and test on real images.

ISOD faces challenges in two areas: the quality of synthetic datasets and the suitability of detection models. Early synthetic image generation \cite{Cutpastelearn,domainrRndomization目标检测,3D1} relied on digital processing or augmentation of real images. These methods depend heavily on real image datasets and require auxiliary models for annotation, increasing complexity, limiting diversity, and potentially producing unrealistic images \cite{diffusionengine}. The emergence of text-to-image generation technologies \cite{stablediffusion,flux,dalle2} has improved the scale and quality of synthetic datasets. Building on these technologies, recent approaches \cite{instagen,ControllableDiffusionModels,Box-to-Image} combine real datasets with diffusion models to synthesize large-scale data. However, these methods employ complex pipelines with multiple hand-designed components and still rely on real datasets, which conflicts with the ISOD \cite{imaginarynet} setting.

\begin{figure*}[!t]
    \centering
    \includegraphics[width=\columnwidth]{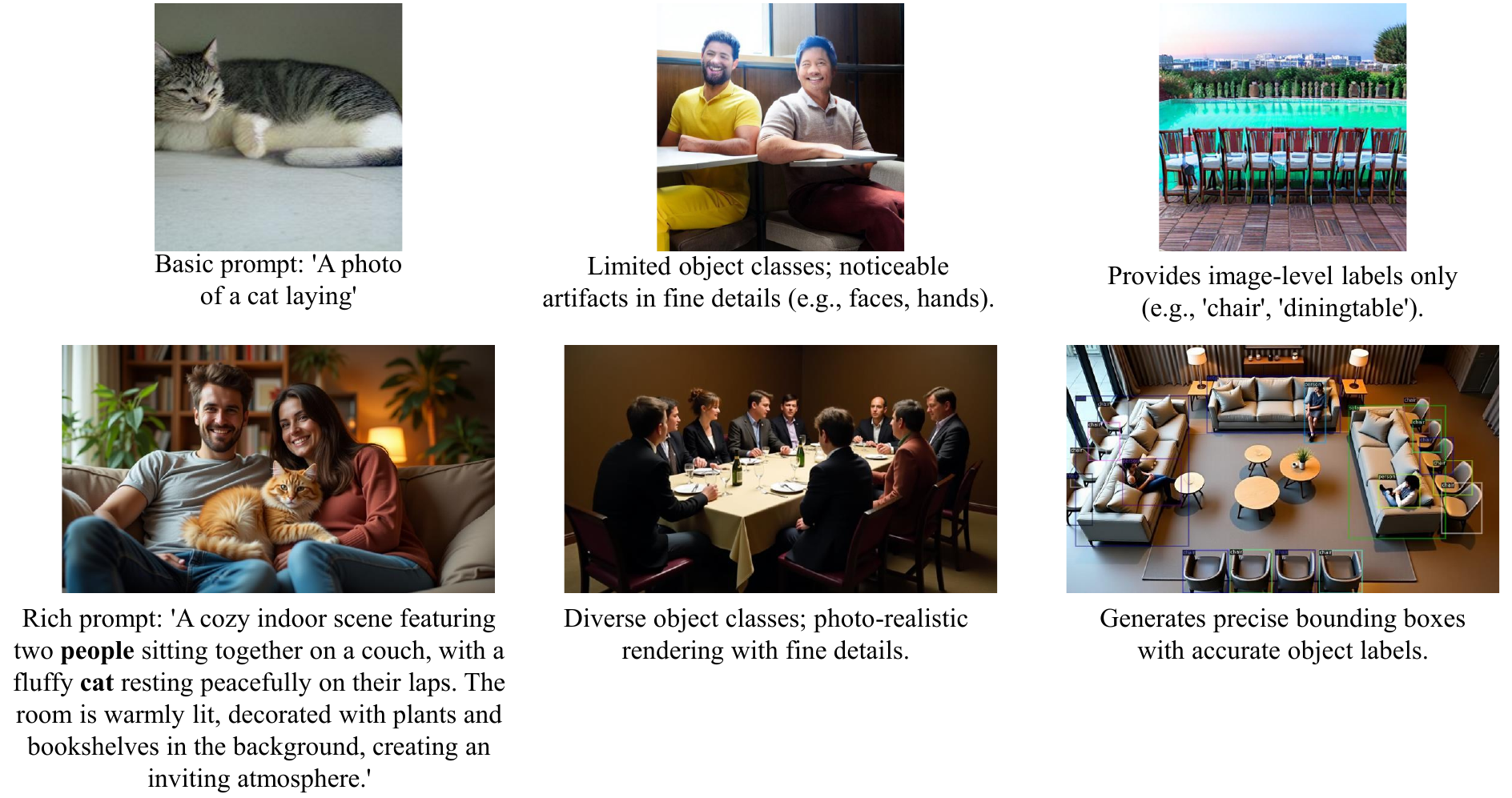}
    \vspace{-1.5em}
    \caption{
        Qualitative comparison of our FluxVOC (bottom) and the baseline \textit{ImaginaryNet} \cite{imaginarynet} (top). Our pipeline achieves superior results in prompt richness, visual fidelity, and annotation granularity.
    }
    \label{fig:dataset_comparison}
    \vspace{-1.5em}
\end{figure*}

ImaginaryNet \cite{imaginarynet} introduced ISOD using GPT-2 \cite{gpt2}-generated prompts for DALL-E 2 \cite{dalle2}, with prompt categories serving as image-level labels for weakly supervised detection. However, as shown in Fig.\ref{fig:dataset_comparison}(top), its synthetic dataset exhibits three key limitations. First, prompts are simplistic, using basic templates, resulting in monotonous scenes with typically one category per image. Second, image quality is constrained by the capabilities of contemporary text-to-image models \cite{dalle2}, with most synthetic images exhibiting obvious detail distortions that violate real-world logic. Third, it provides only image-level labels, limiting localization accuracy. To address these, we first employ LLaMA-3 \cite{llama3} with expert prompts that incorporate inter-category relationships to generate scene-rich and logically consistent descriptions. Second, we leverage the open-source text-to-image model Flux \cite{flux} to generate images with realistic details and high diversity. Third, we employ an open-vocabulary detector Grounding DINO \cite{groundingdino} to automatically generate precise bounding box annotations based on category labels, enabling ISOD to transition from weak to full supervision, as shown in Fig.\ref{fig:dataset_comparison}(bottom).

Regarding model suitability, DETR-based models \cite{detr,deformabledetr,dndetr,dino} achieve strong performance. These models comprise a CNN backbone, an encoder-decoder Transformer \cite{transformer}, and a Hungarian matching module. DAB-DETR \cite{dabdetr} defines object query as a 4D anchor box combined with a content embedding. However, in DAB-DETR and other variants \cite{dndetr,dino,stableDino}, anchor boxes and content embeddings are randomly initialized image-agnostically, which slows convergence and poses a critical challenge in the ISOD setting: models tend to overfit synthetic-specific patterns (e.g., fixed backgrounds and layouts), degrading performance on real images due to the domain gap.

To address this, we propose High-Quality Proposal(HQP)-guided query encoding that uses proposals to initialize anchor boxes and RoI-pooled features to initialize content embeddings. This provides image-specific geometric and semantic priors. Traditional methods \cite{ss,edgebox,bing} generate thousands of proposals per image, increasing computational cost and misleading models toward irrelevant regions. Deep learning approaches \cite{hypernet,proposal2} require training on real images, violating the ISOD setting. In contrast, we employ SAM \cite{SAM} to generate high-quality proposals for synthetic and real images. Since SAM \cite{SAM} proposals cover foreground objects across domains, the model learns domain-consistent low-level visual features (e.g., contours, edges) rather than synthetic-specific patterns (e.g., fixed backgrounds), encouraging generalization.

Annotations in ISOD datasets come from automatic tools \cite{groundingdino}, inevitably containing noise. While DN-DETR \cite{dndetr}'s denoising mechanism accelerates convergence, it creates a problem in ISOD: uniform training pressure causes models to indiscriminately fit error patterns in pseudo-labels. We address this via a key insight: under increasing quality requirements, models face two optimization paths: either fit the high-variance, unstructured noise in pseudo-labels, which becomes difficult under strict standards, or learn stable low-level features like object contours that exhibit high consistency across data, which remains feasible and robust. Inspired by Cascade R-CNN \cite{cascadercnn}, we propose a cascade denoising mechanism that sets increasing IoU thresholds across layers and dynamically adjusts training weights for denoising queries. This guides the model to refine high-quality predictions in later stages, encouraging it to use reliable visual cues from the image to calibrate supervision rather than blindly fitting noisy labels.

This work makes four main contributions:

1) High-quality synthetic data generation pipeline. We design a pipeline using LLaMA-3 \cite{llama3}, Flux \cite{flux}, and Grounding DINO \cite{groundingdino} to construct FluxVOC and FluxCOCO, enabling ISOD to transition from weak to full supervision with superior image quality, scene diversity, and annotation accuracy compared to ImaginaryNet \cite{imaginarynet}.

2) HQP-guided query encoding. We initialize anchor box positions using SAM \cite{SAM}-generated proposals and content embeddings using RoI-pooled features. This provides the model with image-specific priors and encourages learning domain-consistent low-level visual features, improving convergence speed and generalization performance.

3) Cascade denoising algorithm. By dynamically adjusting training weights through progressively increasing IoU thresholds, we guide the model to learn stable low-level visual features rather than fitting noisy labels, mitigating the impact of pseudo-label noise and improving localization accuracy.

4) SOTA performance with high training efficiency. Cascade HQP-DETR achieves 61.04\% mAP@0.5 on the VOC 2007 test set with 12 epochs when trained solely on FluxVOC, surpassing baseline models trained for 30 epochs. It also achieves strong performance on real data, validating the applicability of architectural improvements.

\section{Related Work}
\subsection{Synthetic Images in Object Detection}
In object detection, early synthetic image generation methods relied on digital processing or augmentation of real images. Cut-and-paste methods \cite{Cutpastelearn,cut-paste2} extract foreground objects from real images and composite them onto different backgrounds, domain randomization \cite{domainrRndomization目标检测,domainrRndomization2} vary image parameters to create new samples, and 3D rendering \cite{3D1} combine 3D models with scene information to synthesize datasets. However, these approaches depend on real datasets, require auxiliary models for annotation, and often produce images with limited diversity or unrealistic appearance \cite{diffusionengine}.

Text-to-image generation technologies, including GANs \cite{semantictti} and diffusion models \cite{stablediffusion,flux,dalle2}, enable synthetic datasets with improved scale and quality. InstaGen \cite{instagen} fine-tunes diffusion models on detection datasets to generate annotated datasets; Controllable diffusion models \cite{ControllableDiffusionModels} use visual priors from real images to guide synthesis, reusing original annotations; DiffusionEngine \cite{diffusionengine} combines real images and simulates diffusion noise-denoising processes to obtain annotated images; ODGEN \cite{Box-to-Image} fine-tunes diffusion models on complete images and cropped foregrounds for domain-specific generation. Beyond Generation \cite{Beyondgeneration} uses text-to-image models to generate foreground and background separately, incorporating real context images. Despite advances, all methods depend on real images, conflicting with the ISOD setting.

ImaginaryNet \cite{imaginarynet} introduced the ISOD task, using GPT-2 \cite{gpt2}-generated prompts as input to DALL-E 2 \cite{dalle2} for image synthesis and training weakly supervised detectors with image-level labels. However, their synthetic images suffer from poor detail quality and weak positional supervision, achieving only 70\% of the weakly supervised baseline performance. Our pipeline extends this direction by employing LLaMA-3 \cite{llama3}, Flux \cite{flux}, and Grounding DINO \cite{groundingdino}, enabling high-quality image generation with precise bounding box annotations without relying on real datasets.

\subsection{Object Detection Network Architectures}
Detection models applicable to ISOD include weakly supervised and fully supervised approaches. Weakly supervised methods \cite{onepoint,CLIP-WSDDN} use image-level labels, limiting their localization precision compared to methods trained with box annotations.

For fully supervised detection, CNN-based architectures \cite{resnet} evolved in two directions. Two-stage methods, represented by the R-CNN series \cite{fasterrcnn,cascadercnn},generate region proposals and then perform classification and regression. Proposal quality is critical. Early proposal techniques \cite{ss,bing,edgebox} relied on hand-crafted heuristics and low-level features, but often produced thousands of redundant proposals with incomplete coverage \cite{proposal2}. Deep learning methods \cite{hypernet,proposal2} improved precision but require domain-specific training on real images, conflicting with ISOD's zero-real-data constraint. We leverage SAM \cite{SAM} to generate high-quality proposals for initializing queries, achieving high recall with low redundancy without requiring domain-specific fine-tuning. Single-stage methods like YOLO series \cite{yolo9,yolopx} and SSD \cite{ssd} directly predict boxes and classes in one pass. Despite their efficiency, both paradigms rely on hand-crafted components (e.g., anchor design, NMS), limiting end-to-end optimization flexibility.

Transformers \cite{transformer} achieve strong performance in computer vision \cite{vit, dinov3}. In object detection, DETR \cite{detr} reformulated the task as a set prediction problem with bipartite matching, reducing reliance on hand-designed components like anchors and NMS. Despite this innovation, DETR \cite{detr} suffers from slow convergence and high memory overhead. Subsequent works explored improvements from different angles: works like \cite{deformabledetr,conditionaldetr,SAM-DETR} refine the attention mechanisms, while works like \cite{anchordetr,dabdetr} enhance learning by redefining object queries, with DAB-DETR \cite{dabdetr} defining each object query as a 4D anchor box $(x,y,w,h)$ combined with learnable content embedding.

However, object queries in these models are randomly initialized image-agnostically, forcing the model to learn their distributions purely from training data. Under ISOD settings, this causes overfitting to synthetic-specific patterns like fixed backgrounds. We initialize queries using SAM \cite{SAM}-generated proposals, providing image-specific priors that encourage learning transferable features.

Other methods \cite{dndetr,dino,Do-detr} introduce denoising queries by adding noise to GT boxes as additional supervision. Stable DINO \cite{stableDino} further stabilizes training by incorporating localization into matching and losses. However, under ISOD \cite{imaginarynet} settings, pseudo-labels are inevitably noisy, and uniform treatment of denoising queries causes overfitting. Our cascade denoising addresses this by dynamically weighting queries based on prediction quality, promoting stable learning over fitting noise.

\section{Methodology}
\subsection{Preliminary Knowledge}

DETR \cite{detr} formulates object detection as a set prediction problem, where its Transformer decoder receives learnable object queries that interact with encoder-generated image features to predict bounding boxes and class labels. DETR \cite{detr} employs the Hungarian algorithm for one-to-one matching between predictions and ground truth, then optimizes the model end-to-end using classification loss, L1 loss, and GIoU loss.

DAB-DETR \cite{dabdetr} defines each object query as a combination of a 4D anchor box $A_q$ and a content embedding $C_q$, where $A_q=(x_q, y_q, w_q, h_q)$ is mapped to a high-dimensional positional query $P_q$ via a positional encoding function (PE) and an MLP:
\begin{equation}
    P_q = \text{MLP}(\text{Cat}(\text{PE}(x_q), \text{PE}(y_q), \text{PE}(w_q), \text{PE}(h_q))).
\end{equation}
$C_q$ is a learnable vector that encodes semantic information about object content. At each decoder layer, a prediction head outputs offsets, and object queries are updated.

DN-DETR \cite{dndetr} feeds both denoising queries (derived from GT boxes with added random noise) and regular matching queries into the Transformer decoder. The model performs an additional denoising task to reconstruct the original GT boxes corresponding to the denoising queries. To prevent information leakage between the two query groups, it uses attention masks to isolate them during attention computation.

Stable-DINO \cite{stableDino} modifies the classification loss for positive samples in DETR models. Instead of binary hard labels (0, 1), it uses a positional metric $f_1(s_i)$ that enables the classification score to reflect localization quality:
\begin{equation}
\mathcal{L}_{cls}^{(new)} = \sum_{i=1}^{N_{pos}} ( |f_1(s_i) - p_i|^\gamma \text{BCE}(p_i, f_1(s_i)) ) + \sum_{i=1}^{N_{neg}} p_i^\gamma \text{BCE}(p_i, 0),
\label{eq:stable_dino_cls_loss}
\end{equation}
where $s_i$ is the IoU between the predicted box and the GT box, and $p_i$ is the predicted classification probability. During matching, the classification score is modulated by a positional metric $f_2(s_i')$ to align classification and localization tasks:
\begin{equation}
\mathcal{C}_{cls}^{(new)}(i,j) = |1 - p_i f_2(s_i')|^\gamma \text{BCE}(p_i f_2(s_i'), 1) - (p_i f_2(s_i'))^\gamma \text{BCE}(1 - p_i f_2(s_i'), 1),
\label{eq:stable_dino_cls_match}
\end{equation}
where $s_i'$ is a rescaled GIoU value. To accelerate convergence, Stable-DINO \cite{stableDino} introduces Dense Fusion, which concatenates encoder features with backbone features, projects them to the original dims, and feeds them into the decoder.

\subsection{Overview}
\begin{figure*}[!t]
    \centering
    \includegraphics[width=\textwidth]{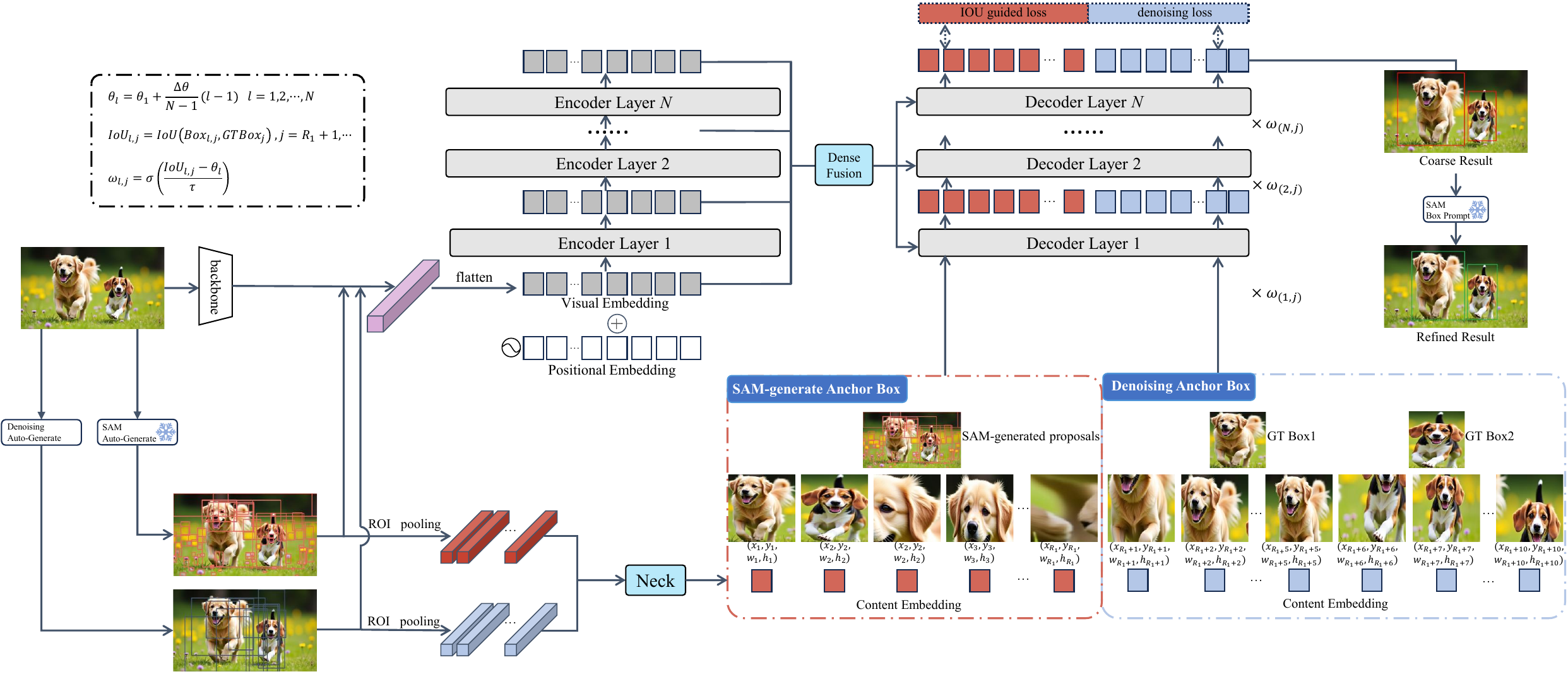}
    \vspace{-2.5em}
    \caption{Overview of Cascade HQP-DETR. Backbone + Encoder (with Dense Fusion) provide features to a decoder processing two branches: the HQP branch initializes object queries with SAM proposals as anchor boxes and RoI-pooled features (via Neck) as content embeddings; the denoising branch initializes denoising queries with noisy GT boxes as anchors and their RoI-pooled features as content embeddings. HQP uses IoU-guided Hungarian matching and IoU-guided classification/regression losses. Denoising applies layer-wise increasing thresholds $\theta_l$, computing $w_{l,j}$ from the IoU between layer-$l$ reconstructed boxes and GT to dynamically modulate features. Inference removes denoising, with predictions optionally refined by SAM.}
    \label{fig:overview}
    \vspace{-1.5em}
\end{figure*}
We propose the Cascade HQP-DETR framework with two components: high-quality synthetic data generation and an enhanced detection architecture.

First, we develop an automated pipeline to construct FluxVOC and FluxCOCO datasets with rich scene diversity and precise annotations. The pipeline employs LLaMA-3 \cite{llama3} for prompt enrichment, Flux \cite{flux} for high-fidelity image synthesis, and Grounding DINO \cite{groundingdino} for automatic bounding box generation (Section~\ref{High-Quality Synthetic Data Generation}).

Second, we enhance the DETR architecture with two innovations (Fig.~\ref{fig:overview}). HQP-guided query encoding employs SAM-generated proposals for anchor box positions and RoI-pooled features for content embeddings, replacing random initialization (Section~\ref{High-Quality Proposal Encoding}). Cascade Denoising sets increasing IoU thresholds across decoder to dynamically weight denoising queries (Section~\ref{Cascade Denoising Algorithm}). We integrate IoU-guided classification loss, position-modulated matching, and Dense Fusion from Stable-DINO \cite{stableDino}.

\subsection{High-Quality Synthetic Data Generation}
\label{High-Quality Synthetic Data Generation}
We construct FluxVOC and FluxCOCO datasets through a three-stage pipeline: prompt generation, image synthesis, and automatic annotation.

\subsubsection{Prompt Generation}
Given a target benchmark dataset (PASCAL VOC \cite{pascalvoc}, COCO \cite{coco}) with $C$ object categories $\mathcal{C}=\{c_1, c_2, \cdots, c_C\}$, we partition the category set into three frequency-based tiers: $\mathcal{C}_{common}$, $\mathcal{C}_{moderate}$, and $\mathcal{C}_{rare}$. We assign sampling weights $\{w_{common}, w_{moderate}, w_{rare}\}$ with $w_{common} > w_{moderate} > w_{rare}$ to reflect real-world object frequency distributions. For each sample, we randomly draw $n_{obj} \sim \text{Uniform}(1, 4)$ categories from $\mathcal{C}$ according to their weights, forming a category combination $\mathcal{C}_{sample}=\{c_{s_1}, \cdots, c_{s_{n_{obj}}}\}$. This combination is converted into a basic prompt template $T_{\mathrm{basic}}$:
\begin{quote}
    \textit{``A realistic photo which contains $c_{s_1}$, $c_{s_2}$, $\cdots$, and $c_{s_{n_{obj}}}$.''}
\end{quote}

To enrich scene descriptions with vivid details, context, and plausible instance counts, we employ the large language model LLaMA-3 \cite{llama3} via an expert prompt $\mathcal{I}_{sys}$:
\begin{tcolorbox}[
    colback=Gray,         
    colframe=darkgray!75,
    arc=2mm,             
    boxsep=5pt,           
    left=10pt,            
    right=10pt            
]
\begin{list}{}{\setlength{\leftmargin}{0pt} \setlength{\rightmargin}{0pt}}
\item[] \textit{You are an expert in generating high-quality prompts for text-to-image models. Your task is to take a basic prompt and enhance it by adding vivid details, specific attributes, context, lighting, and atmosphere. Also, for each object mentioned, determine and assign a \textbf{plausible number of instances that would naturally fit the scene you are creating. The count should be guided by realism and the typical context of the objects.} The final output must be a single, concise, highly descriptive prompt in English.}\\
\textit{Basic Prompt: $T_{\mathrm{basic}}$}
\end{list}
\end{tcolorbox}

This produces an enriched prompt $P_{\mathrm{rich}}$. For example, \textit{``a realistic photo which contains person and dog''} is transformed into \textit{``A heartwarming outdoor photo of a young woman in casual clothing playing in a sunny park with her two golden retrievers, while autumn leaves gently fall around them under soft afternoon sunlight``}.

\subsubsection{Image Synthesis.}
The enriched prompt $P_{\mathrm{rich}}$ is fed into Flux \cite{flux}, a text-to-image diffusion model, to synthesize high-fidelity images $I_{\mathrm{syn}}$.

\subsubsection{Automatic Annotation.}
To obtain instance-level bounding box annotations, we employ the open-vocabulary detector Grounding DINO \cite{groundingdino}. We construct a detection prompt $P_{\mathrm{det}}$ directly from the category names in the basic template. For example, if the basic template is \textit{a realistic photo of a cat and a dog}, then $P_{\mathrm{det}}$ is \textit{cat. dog.} Applying Grounding DINO \cite{groundingdino} to $(I_{\mathrm{syn}}, P_{\mathrm{det}})$ yields annotations $A_{\mathrm{syn}}$ containing bounding boxes and class labels.

\begin{figure*}[!t]
    \centering
    \vspace{-2.8em}
    \includegraphics[width=0.95\textwidth]{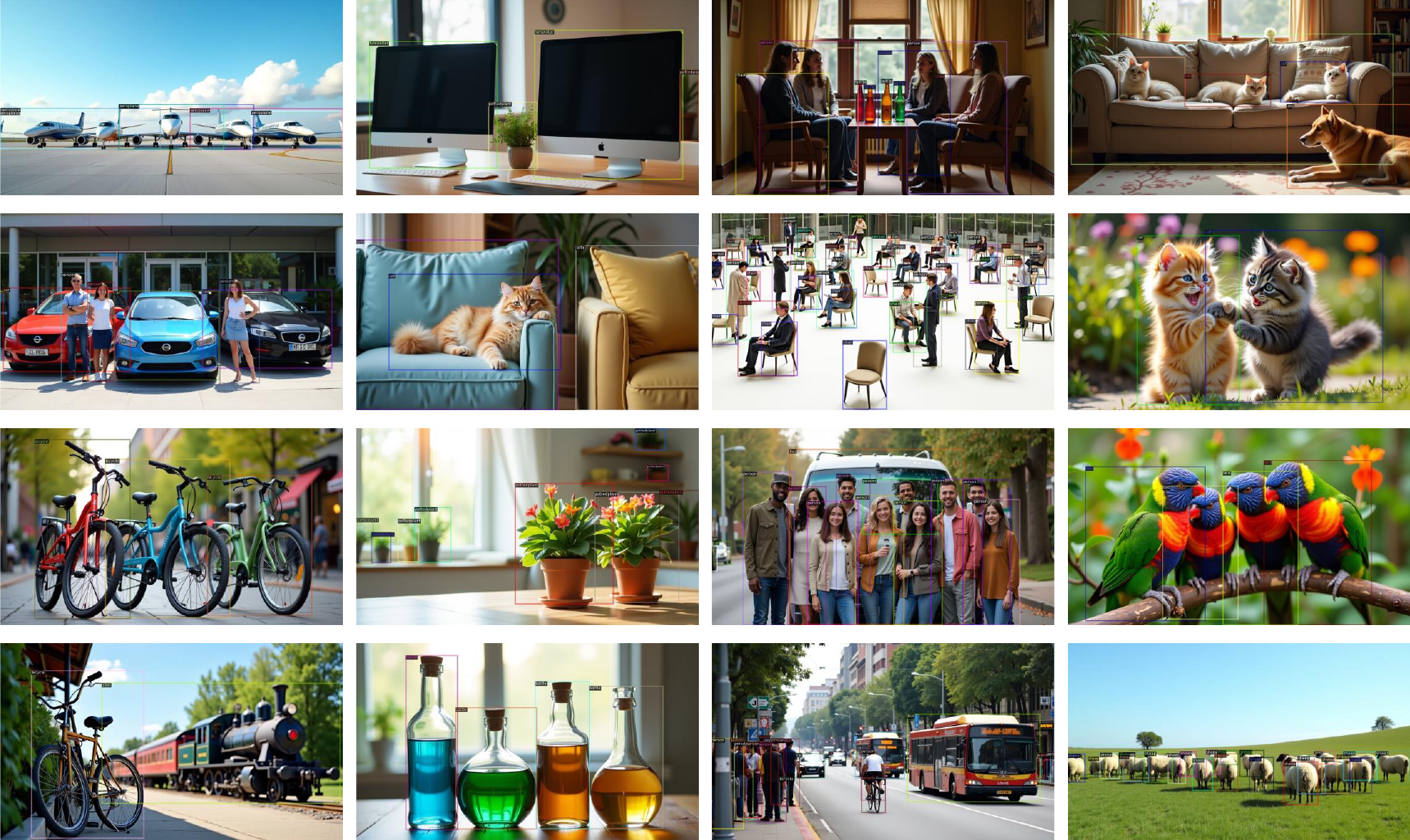} 
    \vspace{-0.6em}
    \caption{Visual examples from our synthetically generated FluxVOC dataset, shown with their automated bounding box annotations. The collage is thematically organized by column to showcase the dataset's breadth. From left to right, the columns display: (1) various modes of transportation, (2) common indoor objects, (3) complex, multi-instance human-centric scenes, and (4) a diverse range of animals. This structure highlights our pipeline's ability to generate high-quality images across distinct domains, including challenging scenarios with small, dense, and occluded objects. (Best viewed in color and with zoom).}
    \label{fig:fluxvoc}
    \vspace{-1em}
\end{figure*}

Fig.~\ref{fig:fluxvoc} showcases samples from FluxVOC across various domains, including transportation, indoor objects, multi-instance human-centric scenes, and animals. The pipeline generates challenging scenarios like small, dense, and occluded objects, providing comprehensive training data for ISOD.
Crucially, this synthetic-plus-pseudo-label pipeline provides fully supervised training signals with zero manual annotation, showing that high-fidelity virtual images paired with reliable pseudo boxes are sufficient to train competitive detectors while dramatically reducing data-collection and labeling cost.

\subsection{High-Quality Proposal-Guided Query Encoding}
\label{High-Quality Proposal Encoding}
DAB-DETR \cite{dabdetr} represents each object query as a combination of a 4D anchor box and a content embedding. However, both components are randomly initialized without input image information, which we term the double-blind initialization problem. This poses two challenges in ISOD: (1) the model must learn query distributions from scratch, slowing convergence; (2) models tend to overfit synthetic-specific patterns like background styles and object layouts, degrading real-world generalization. To address this, we propose HQP encoding providing image-specific geometric and semantic priors through two mechanisms described below.

\subsubsection{Geometry-Aware Priors from SAM Proposals}
We use SAM \cite{SAM} to generate high-quality region proposals. As a foundation model trained on large-scale data, SAM \cite{SAM} produces accurate, class-agnostic segmentation masks without domain-specific fine-tuning. We employ SAM's automatic mask generation mode to produce class-agnostic segmentation masks. For each mask $m_i$, we compute its minimum enclosing bounding box:
\begin{equation}
\mathcal{P}_{SAM} = \{p_1, p_2, ..., p_{N_p}\}, \quad p_i = (x_i, y_i, w_i, h_i) = \text{BBox}(m_i).
\end{equation}

These proposals have three properties: (1) \textbf{high recall} through multi-scale sampling that ensures comprehensive coverage of salient objects; (2) \textbf{low redundancy} with only $N_p$ proposals (typically 150-200) retained per image, compared to thousands in traditional methods \cite{ss,edgebox,bing,fasterrcnn}; (3) \textbf{domain consistency} as SAM \cite{SAM} accurately covers foreground objects in both synthetic and real domains.

These proposals $\mathcal{P}_{SAM}$ directly initialize the anchor boxes of our object queries. Compared to DAB-DETR \cite{dabdetr}'s 300 randomly initialized queries, our approach produces an adaptive number of queries focused on high-potential regions from the outset.

\subsubsection{Semantic-Aware Priors from Region Features}
To complement geometric priors with semantic information, we extract region features from the backbone feature map to initialize content embeddings.

Given an input image $I$, a CNN backbone (e.g., ResNet-50) produces the feature map $F_{C5} \in \mathbb{R}^{C \times H' \times W'}$ from its final convolutional layer, where $C$ is the channel dimension. For each proposal $p_i$, we apply RoI Pooling \cite{fastrcnn}:
\begin{equation}
    f_i^{\text{roi}} = \text{RoI Pooling}(F_{C5}, p_i) \in \mathbb{R}^{C \times H_{\text{roi}} \times W_{\text{roi}}}.
\end{equation}
which extracts a fixed-size feature by spatial pooling over the corresponding region.

The extracted feature is projected to the decoder dimension via a Neck module:
\begin{equation}
    c_i^{\text{init}} = \text{Neck}_{\theta}(f_i^{\text{roi}}),
\end{equation}
where $\text{Neck}_{\theta}$ comprises two linear layers with ReLU activations: $\text{Linear}_1: \mathbb{R}^{C \times H_{\text{roi}} \times W_{\text{roi}}} \rightarrow \mathbb{R}^{d_{\text{hidden}}}$ and $\text{Linear}_2: \mathbb{R}^{d_{\text{hidden}}} \rightarrow \mathbb{R}^{d_{\text{model}}}$, with $d_{\text{model}}=256$.

This provides each content embedding $c_i^{\text{init}}$ with semantic information from its corresponding region.

\subsubsection{Synergy of Geometric and Semantic Priors}
Geometric and semantic priors work synergistically. Geometric priors focus feature extraction on high-potential regions, while semantic priors provide initial understanding. Together, they transform the decoder's task from global exploration to local refinement and classification, simplifying optimization.

\subsection{Cascade Denoising Algorithm}
\label{Cascade Denoising Algorithm}
DN-DETR \cite{dndetr} accelerates DETR convergence through a denoising training task, but applies uniform training pressure to all denoising queries. This uniform strategy becomes unsuitable in the ISOD setting, where pseudo-labels generated by automatic annotation tools \cite{groundingdino} inevitably contain noise. To address this, we propose the Cascade Denoising algorithm inspired by Cascade R-CNN \cite{cascadercnn}. The algorithm dynamically adjusts training weights through progressively increasing IoU thresholds across decoder layers. It consists of two key components described below.

\subsubsection{Dynamic IoU Threshold Design}
We define an increasing threshold $\theta_l$ for each decoder layer $l$:
\begin{equation}
    \theta_l = \theta_1 + \frac{\Delta\theta}{N-1}(l-1), \quad l \in \{1, 2, \ldots, N\},
\end{equation}
where $\theta_1=0.3$ is the initial threshold, $\Delta\theta=0.6$ is the total increment, and $N=6$ is the total number of decoder layers. The threshold increases from 0.3 in the first layer to 0.9 in the final layer.

\subsubsection{Query Weighting Mechanism}
Based on the threshold $\theta_l$, we compute a dynamic weight for each denoising query $j$ at layer $l$. At each decoder layer, we obtain box predictions from denoising queries. We calculate the IoU between the predicted box $Box_{l,j}$ at layer $l$ and its corresponding ground truth box $GTBox_j$:
\begin{equation}
    IoU_{l,j} = \text{IoU}(Box_{l,j}, GTBox_j).
\end{equation}
and then map the deviation of $IoU_{l,j}$ relative to threshold $\theta_l$ into a training weight via a sigmoid function:
\begin{equation}
    \omega_{l,j} = \sigma\left(\frac{IoU_{l,j} - \theta_l}{\tau}\right),
\end{equation}

where $\sigma(x) = 1/(1+\exp(-x))$ is the sigmoid function, and the temperature $\tau=0.1$ controls the steepness of the weighting curve.
This weight directly modulates the feature of the denoising query:
\begin{equation}
    f'_{dn,j} = \omega_{l,j} \cdot f_{dn,j}
\end{equation}
The modulated feature $f'_{dn,j}$ is used to compute the denoising loss. This creates three training scenarios:
\begin{itemize}
    \item \textbf{High-quality}: When $IoU_{l,j} \gg \theta_l$, $\omega_{l,j} \approx 1$, the query receives full pressure.
    \item \textbf{Low-quality}: When $IoU_{l,j} \ll \theta_l$, $\omega_{l,j} \approx 0$, the training signal is suppressed.
    \item \textbf{Boundary}: When $IoU_{l,j} \approx \theta_l$, $\omega_{l,j} \approx 0.5$.
\end{itemize}
This enables early layers to accept queries with lower IoU for training, while later layers focus on queries with higher IoU. Since each decoder layer reconstructs boxes from its denoising queries, this creates a progressive refinement process where the model learns coarse localization first and specializes in precise boundary refinement.

\subsection{Training and Inference Pipeline}
\subsubsection{Training Pipeline}
Our Cascade HQP-DETR model is trained end-to-end on synthetic images. We adopt two strategies from Stable-DINO \cite{stableDino}: (1) \textbf{Dense Fusion} concatenates encoder features with backbone features and projects them to the decoder dims; (2) \textbf{Stable Matching} employs position-modulated matching cost and position-supervised classification loss to align classification and localization.

The training proceeds as follows: An image passes through a CNN backbone and a Transformer encoder (integrated with Dense Fusion) to generate context-rich feature maps. The Transformer decoder then processes two groups of queries in parallel, isolated by an attention mask: (1) object queries initialized with our High-Quality Proposals; (2) denoising queries progressively refined through our Cascade Denoising algorithm. Denoising queries are generated by adding noise to GT boxes, and their features are modulated via our proposed dynamic IoU-guided weighting mechanism.

The total loss function is a weighted sum of three components:
\begin{equation}
    L_{total} = \lambda_{cls}\mathcal{L}_{cls} + \lambda_{box}\mathcal{L}_{box} + \lambda_{dn}\mathcal{L}_{dn},
\end{equation}
where $\mathcal{L}_{cls}$ is the IoU-guided classification loss, $\mathcal{L}_{box}$ is the bounding box regression loss (combining L1 and GIoU), and $\mathcal{L}_{dn}$ is the denoising loss. Hungarian matching employs a position-modulated strategy when computing the matching cost.

\subsubsection{Inference Pipeline}
During inference, the denoising branch is removed, and only object queries initialized with HQP generate predictions. After the image passes through backbone and encoder, SAM-generated proposals initialize anchor boxes, RoI-pooled features initialize content embeddings, and decoder refines these queries to output bounding boxes and class predictions. Optionally, SAM refines predicted boxes via segmentation masks.

\section{Why Our Approach is Effective for ISOD}
\subsection{HQP-Guided Query Encoding for Cross-Domain Generalization}
\begin{figure}[t]
    \centering
    \includegraphics[width=\columnwidth]{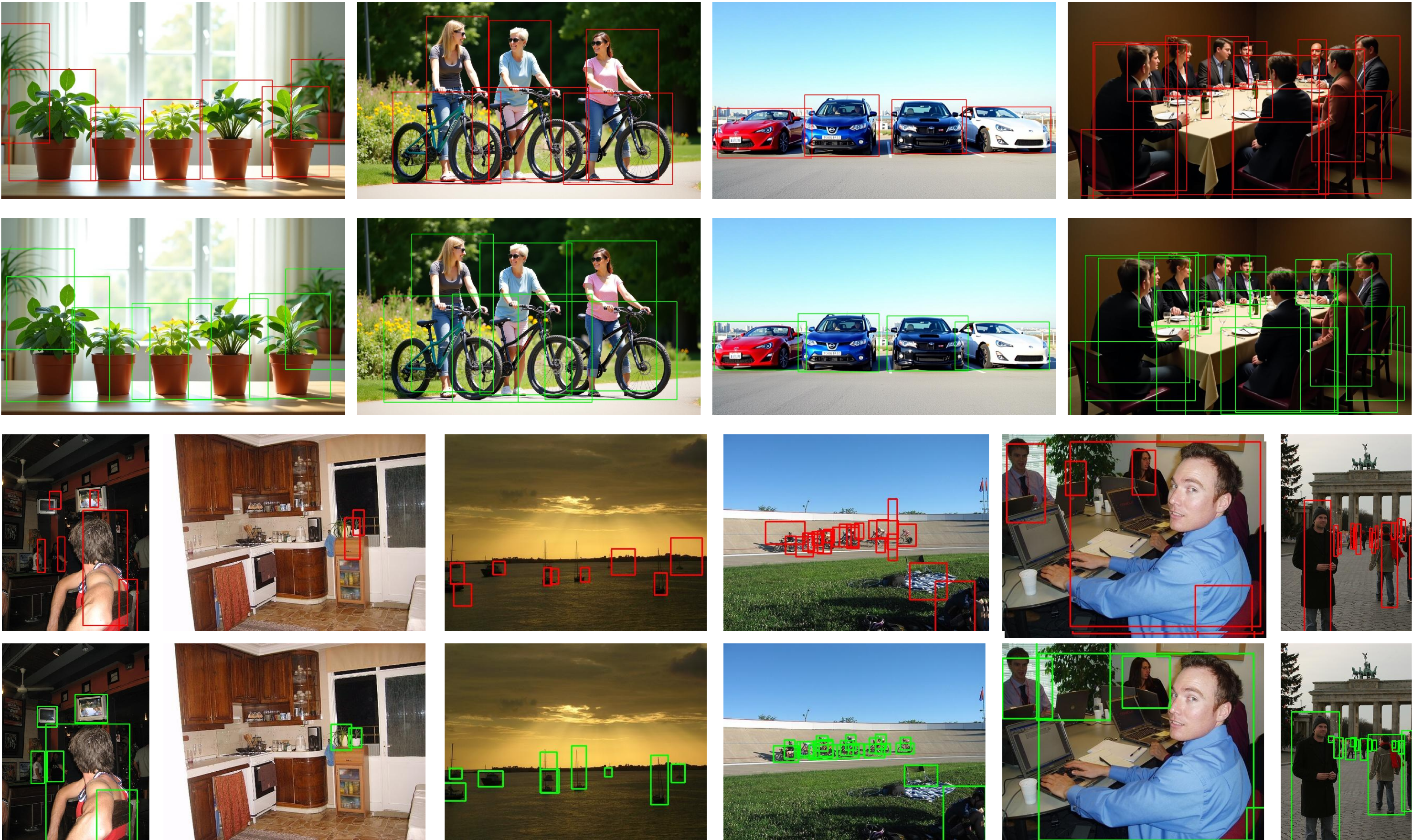}
    \caption{
        First-layer matched queries for DAB-DETR \cite{dabdetr} (red) vs. our Cascade HQP-DETR (green).\textbf{Synthetic Domain (top two rows):} On the FluxVOC training data, both models exhibit high initial IoU, demonstrating their ability to fit the synthetic distribution.\textbf{Real Domain (bottom two rows):} On the real VOC07 \cite{pascalvoc} test set, our model's queries remain highly accurate, while the baseline's degrade significantly. This highlights our model's superior generalization across the synthetic-to-real domain gap.
    }
    \label{fig:objectquery}
    \vspace{-1.1em}
\end{figure}
A key challenge in ISOD is the significant distribution discrepancy between synthetic and real domains. Mainstream DETR variants (\cite{dabdetr,dndetr,dino}) initialize object queries image-agnostically.When trained exclusively on synthetic data, these queries inevitably overfit to synthetic-specific visual patterns (e.g., stereotyped background styles, predictable object layouts), resulting in poor generalization to real images.

Our HQP mechanism addresses this overfitting issue by leveraging SAM \cite{SAM}-generated proposals and corresponding RoI-pooled features for initialization. This endows each query with image-specific geometric and semantic priors, solving the crucial initial problem of "where to look." Consequently, the decoder's learning task is greatly simplified, transforming from global exploration to focused local refinement: precisely fine-tuning the proposals to match object contours and classifying these features.

More importantly, this design redirects the model's learning focus. It steers the model away from the synthetic domain's non-transferable high-level contextual patterns, guiding it instead toward highly transferable low-level visual features like object contours and component structures. Local contours of human limbs or dog tails, for instance, exhibit strong visual consistency across both synthetic and real domains, thus providing a robust foundation for generalization.

Figure~\ref{fig:objectquery} provides intuitive validation by visualizing queries from the first decoder layer that matched the GT boxes. In the synthetic domain (top two rows), these queries demonstrate accurate target localization for both models, proving their ability to fit the training distribution. However, on real VOC 2007 images (bottom two rows), our model's matched queries remain highly accurate, while the baseline's queries become scattered and poorly aligned.

\subsection{Cascade Denoising for Noisy Pseudo-Labels}
Another key challenge in ISOD is that annotations are derived from automated annotation tools \cite{groundingdino}, which inevitably introduce systematic noise into pseudo-labels. DN-DETR \cite{dndetr} applies uniform training pressure to all denoising queries, causing the model to indiscriminately fit both accurate and erroneous supervision signals.

Our Cascade Denoising mechanism dynamically adjusts training weights based on prediction quality. Instead of uniform pressure, we set progressively increasing IoU thresholds across decoder layers, allowing denoising queries with large initial errors to participate in early learning while focusing on high-quality predictions in later stages. Under this increasingly stringent pressure, the optimizer faces two optimization paths: fit the high-variance, unstructured noise within pseudo-labels, or learn stable and consistent visual features of the object itself. The former represents a much more difficult optimization path. In comparison, the physical contours of an object, as a low-level visual feature, exhibit minimal variance and high consistency, providing the optimizer with a simpler and more robust convergence target. Thus, our mechanism encourages leveraging reliable visual features from the image itself rather than uncritically mimicking noisy pseudo-labels, learning more robust representations of object boundaries.

\begin{table}[t]
  \centering
  \footnotesize
  \renewcommand{\arraystretch}{0.8}
  \caption{Robustness to annotation noise on the VOC 2007 test set (mAP@0.5, \%). Models are trained on FluxVOC with simulated noise and evaluated on the clean test set.}
  \label{tab:noise_robustness}
  \begin{tabular}{l S[table-format=2.2] S[table-format=2.2] S[table-format=1.2]}
    \toprule
    \textbf{Noise Level} & {\textbf{Cascade DN}} & {\textbf{Uniform DN}} & {\textbf{Gap}} \\
    \midrule
    0\% (Clean)  & 61.01 & 59.72 & 1.29 \\
    5\%          & 60.79 & 59.47 & 1.32 \\
    10\%         & 59.38 & 57.82 & 1.56 \\
    15\%         & 56.22 & 54.78 & 1.44 \\
    20\%         & 52.57 & 51.10 & 1.47 \\
    \bottomrule
  \end{tabular}
  \vspace{-1.5em}
\end{table}

To validate this robustness claim, we simulate varying levels of annotation noise by adding Gaussian perturbations to the bounding box coordinates in FluxVOC dataset. Specifically, for a box with width $w$ and height $h$, we add noise $\mathcal{N}(0, \sigma^2)$ to each coordinate, where $\sigma = \text{noise\_level} \times \{w, h\}$. We compare Cascade DN against Uniform DN—our model with uniform denoising weights (i.e., $w_l(q_{dn}) = 1$ for all queries, equivalent to DN-DETR's strategy). Table~\ref{tab:noise_robustness} presents results on the clean VOC 2007 test set. As noise increases from 0\% to 20\%, both methods degrade, but Cascade Denoising consistently outperforms Uniform DN. Even under 20\% noise, our method maintains a consistent +1.47 mAP advantage, validating our hypothesis: by progressively down-weighting low-quality predictions, our mechanism guides the model to rely on stable visual features rather than uncritically fitting noisy pseudo-labels.

\section{Experiments}
\subsection{Settings}
\subsubsection{Datasets and Evaluation Metrics}
For ISOD evaluation, we train on two synthetic datasets (Section~\ref{High-Quality Synthetic Data Generation}): FluxVOC (10k images, 20 categories) and FluxCOCO (80k images, 80 categories), comparable to PASCAL VOC 2012 ~\cite{pascalvoc} trainval and MS COCO train2014 ~\cite{coco}. To validate architectural improvements on real data, we also train on PASCAL VOC 2012 trainval.

We evaluate on PASCAL VOC 2007 test~\cite{pascalvoc}, reporting mAP@0.5, AP@[0.5:0.95], and AP@0.75. We additionally test on MS COCO 2014~\cite{coco} val using AP, AP$_{50}$, AP$_{75}$, AP$_S$, AP$_M$, AP$_L$ (see ~\ref{FluxCOCO Performance on the MS COCO Benchmark}).

\subsubsection{Implementation Details}
All experiments use ResNet-50 \cite{resnet} backbone. The encoder and decoder each have 6 layers with $d_{model}=256$. For our proposed components: (1) \textbf{HQP-Guided Query Encoding}: SAM-H \cite{SAM} generates proposals via automatic mask generation (64 points per side), with RoI-pooled $7 \times 7$ features projected to 256-d queries via a two-layer MLP (hidden dim 4,096). (2) \textbf{Cascade Denoising}: IoU threshold $\theta_l$ increases linearly from 0.3 to 0.9 across layers ($\tau=0.1$, box noise 0.4, 5 groups).

We employ Focal Loss \cite{focalloss} ($\alpha=0.25$, $\gamma=2.0$) with weights $(\lambda_{cls}, \lambda_{bbox}, \lambda_{giou})=(6.0, 5.0, 2.0)$. Hungarian matching uses focal cost with coefficients $(2.0, 5.0, 2.0)$ and coupling $\beta=0.5$ \cite{stableDino}.Training uses AdamW(lr $7.5 \times 10^{-5}$, weight decay $1 \times 10^{-4}$, batch size 24) on four A100s for 12 epochs (FluxVOC) and 24 epochs (FluxCOCO). Other settings follow \cite{dndetr,stableDino}.

\subsection{Dataset Quality Evaluation}
To evaluate FluxVOC quality, we compare it with ImaginaryNet \cite{imaginarynet} across multiple dimensions. Since the original ImaginaryNet lacks bounding box annotations, we re-annotate it using CLIP \cite{clip} for class inference and Grounding DINO \cite{groundingdino} for box generation, yielding a fully supervised baseline for comparison.

\begin{table}[!htbp]
  \centering
  \footnotesize
  \renewcommand{\arraystretch}{0.75}
  \caption{Instance count per category comparison. FluxVOC provides 4$\times$ more instances than the re-annotated ImaginaryNet, with better alignment to VOC 2012 distribution.}
  \label{tab:instance_counts}
  \setlength{\tabcolsep}{2.5pt} 
  \begin{tabular}{
    l
    S[table-format=5.0]
    S[table-format=5.0]
    S[table-format=3.0]
    r
    @{\hskip 0.4cm}
    l
    S[table-format=5.0]
    S[table-format=5.0]
    S[table-format=3.0]
    r
  }
    \toprule
    \textbf{Category} & 
    {\textbf{VOC12}} & 
    {\textbf{Flux}} & 
    {\textbf{Img}} &
    {\textbf{Gain}} &
    \textbf{Category} & 
    {\textbf{VOC12}} & 
    {\textbf{Flux}} & 
    {\textbf{Img}} &
    {\textbf{Gain}} \\
    \midrule
    Aeroplane   & 954   & \bfseries 1248 & 514 & \showgain{1248}{514} &
    Diningtable & 747   & \bfseries 1151 & 366 & \showgain{1151}{366} \\
    
    Bicycle     & 790   & \bfseries 1116 & 521 & \showgain{1116}{521} &
    Dog         & 2062  & \bfseries 2514 & 583 & \showgain{2514}{583} \\
    
    Bird        & 1221  & \bfseries 1519 & 575 & \showgain{1519}{575} &
    Horse       & 750   & \bfseries 1420 & 674 & \showgain{1420}{674} \\
    
    Boat        & 999   & \bfseries 1499 & 563 & \showgain{1499}{563} &
    Motorbike   & 751   & \bfseries 1197 & 570 & \showgain{1197}{570} \\
    
    Bottle      & 1482  & \bfseries 1904 & 561 & \showgain{1904}{561} &
    Person      & 10320 & \bfseries 10914& 607 & \showgain{10914}{607} \\
    
    Bus         & 637   & \bfseries 1162 & 502 & \showgain{1162}{502} &
    Pottedplant & 1099  & \bfseries 2603 & 563 & \showgain{2603}{563} \\
    
    Car         & 2364  & \bfseries 2797 & 567 & \showgain{2797}{567} &
    Sheep       & 994   & \bfseries 1374 & 377 & \showgain{1374}{377} \\
    
    Cat         & 1227  & \bfseries 1612 & 608 & \showgain{1612}{608} &
    Sofa        & 786   & \bfseries 1213 & 626 & \showgain{1213}{626} \\
    
    Chair       & 2906  & \bfseries 3344 & 594 & \showgain{3344}{594} &
    Train       & 876   & \bfseries 1398 & 566 & \showgain{1398}{566} \\
    
    Cow         & 1102  & \bfseries 1205 & 628 & \showgain{1205}{628} &
    TV  & 868   & \bfseries 1206 & 276 & \showgain{1206}{276} \\
    
    \midrule
    \multicolumn{10}{c}{\textbf{Total:} VOC 2012 = 33,835 \quad FluxVOC = 44,396 \quad ImaginaryNet = 11,001 \quad \textbf{Gain:} \showgain{44396}{11001}} \\
    \bottomrule
  \end{tabular}
\end{table}
\subsubsection{Instance Quantity and Distribution}
As shown in Table~\ref{tab:instance_counts}, FluxVOC contains 44,396 instances compared to 11,001 in ImaginaryNet \cite{imaginarynet}—a 4$\times$ increase. The improvement is consistent across all 20 categories, with large gaps in \textit{person} (10,914 vs. 607), \textit{pottedplant} (2,603 vs. 563), and \textit{TV} (1,206 vs. 276). More importantly, FluxVOC's distribution better approximates real VOC 2012 \cite{pascalvoc}, which is critical for effective sim-to-real transfer.

\subsubsection{Spatial and Visual Diversity}
\begin{figure}[htbp]
    \centering
    \includegraphics[width=0.9\linewidth]{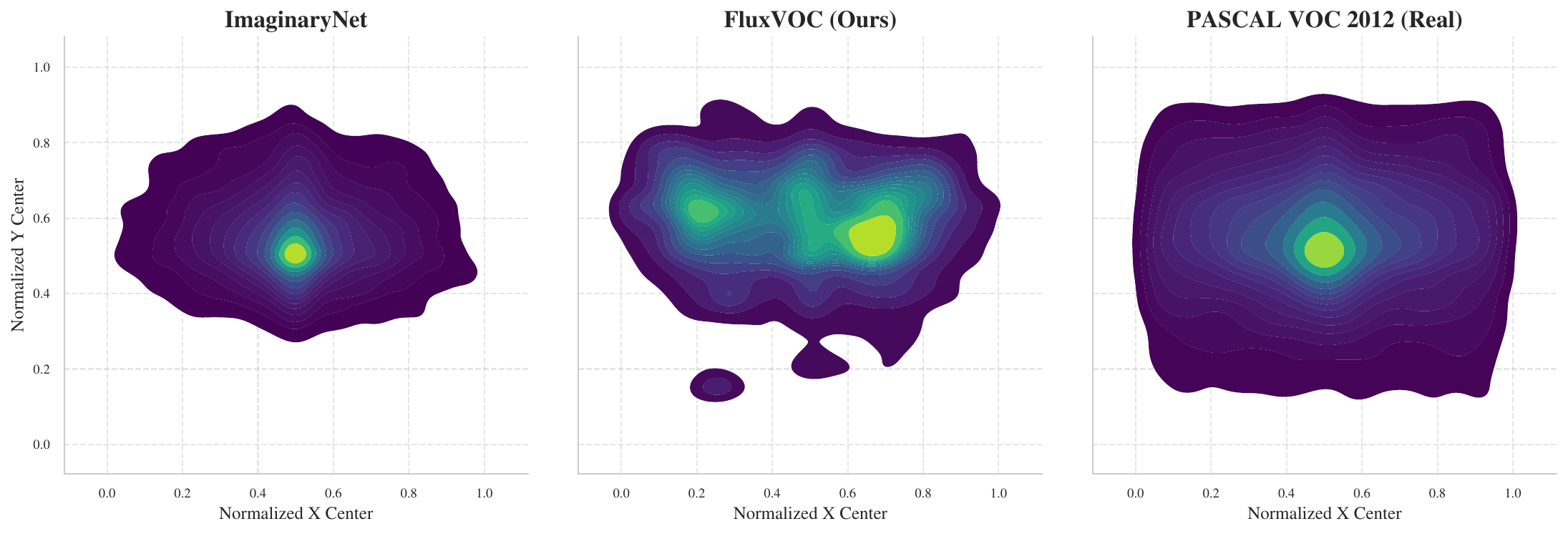}
    \caption{KDE of bounding box centers. ImaginaryNet (left): extreme center bias. FluxVOC (center): broad, multi-modal. Real VOC 2012 (right): natural, broad coverage.}
    \label{fig:spatial_dist}
\end{figure}

We visualize spatial diversity via KDE plots of normalized bounding box centers (Fig.~\ref{fig:spatial_dist}). ImaginaryNet (left) shows extreme center bias with a sharp density peak, placing most objects in a narrow central region—this causes models to overfit positional priors and fail on off-center test objects. Real VOC 2012 (right) exhibits natural central tendency with broad coverage. FluxVOC (center) similarly features a broad distribution, multiple density peaks, and distinct object "islands" in peripheral regions, matching the complexity of real data and encouraging position-invariant learning.

\subsubsection{Performance Evaluation}
To quantify dataset quality, we train DAB-DETR \cite{dabdetr} on different data sources and evaluate on VOC 2007 \cite{pascalvoc} test (Table~\ref{tab:dataset_performance}). In the synthetic-only setting, FluxVOC achieves 44.85\% mAP@0.5, outperforming re-annotated ImaginaryNet \cite{imaginarynet} (35.14\%) by +9.71 points, with consistent gains on stricter metrics (AP@[0.5:0.95]: 22.58\% vs. 19.52\%, AP@0.75: 19.32\% vs. 17.84\%).

When mixed with real VOC 2012 data, the quality difference becomes more pronounced. FluxVOC + VOC12 achieves 74.51\% mAP@0.5, surpassing both ImaginaryNet \cite{imaginarynet} + VOC12 (68.14\%) and the real-data-only baseline (68.16\%) by +6.35 points. Notably, adding ImaginaryNet \cite{imaginarynet} to real data is ineffective, resulting in a slight performance degradation (68.14\% vs. 68.16\%), whereas FluxVOC acts as effective data augmentation, demonstrating that synthetic data quality, not volume, determines its utility for ISOD.
\begin{table*}[!t]
  \centering
  \caption{Performance comparison of DAB-DETR \cite{dabdetr} trained on different data sources, evaluated on VOC 2007 test set \cite{pascalvoc}. FluxVOC significantly outperforms ImaginaryNet in both synthetic-only and mixed-data settings.}
  \label{tab:dataset_performance}
  \sisetup{table-format=2.2}
  \small
  \renewcommand{\arraystretch}{0.75}
  \resizebox{\textwidth}{!}{%
  \begin{tabular}{l S S S S S S S}
    \toprule
    & \multicolumn{2}{c}{\textbf{Synthetic Only}} 
    & \multicolumn{2}{c}{\textbf{Synthetic + Real}} 
    & \multicolumn{2}{c}{\textbf{Real Only}}
    & {\textbf{Baseline}} \\
    \cmidrule(lr){2-3} \cmidrule(lr){4-5} \cmidrule(lr){6-7} \cmidrule(lr){8-8}
    \textbf{Category} 
    & {\textbf{FluxVOC}} & {\textbf{ImgNet$^{\dagger}$}} 
    & {\textbf{FluxVOC + VOC12}} & {\textbf{ImgNet$^{\dagger}$ + VOC12}} 
    & {\textbf{VOC12}} & {\textbf{VOC07+12}} 
    & {\textbf{ImgNet$^{\ddagger}$}} \\
    \midrule
    Aeroplane   & {\showPerfChange{53.67}{55.52}} & 55.52 & {\showPerfChange{76.51}{75.90}} & 75.90 & 75.06 & 81.41 & 47.18 \\
    Bicycle     & {\showPerfChange{61.29}{52.07}} & 52.07 & {\showPerfChange{81.70}{77.54}} & 77.54 & 77.92 & 84.12 & 43.13 \\
    Bird        & {\showPerfChange{44.85}{33.91}} & 33.91 & {\showPerfChange{79.35}{70.74}} & 70.74 & 71.21 & 84.41 & 35.51 \\
    Boat        & {\showPerfChange{33.83}{39.33}} & 39.33 & {\showPerfChange{63.82}{62.04}} & 62.04 & 53.46 & 69.44 & 5.33 \\
    Bottle      & {\showPerfChange{31.47}{15.55}} & 15.55 & {\showPerfChange{54.90}{43.82}} & 43.82 & 46.02 & 65.21 & 19.23 \\
    Bus         & {\showPerfChange{66.44}{41.40}} & 41.40 & {\showPerfChange{83.19}{75.16}} & 75.16 & 74.03 & 81.98 & 32.58 \\
    Car         & {\showPerfChange{63.13}{50.93}} & 50.93 & {\showPerfChange{82.60}{75.42}} & 75.42 & 78.02 & 85.80 & 41.23 \\
    Cat         & {\showPerfChange{57.35}{63.21}} & 63.21 & {\showPerfChange{86.00}{85.69}} & 85.69 & 86.84 & 89.02 & 66.72 \\
    Chair       & {\showPerfChange{28.32}{8.01}}  & 8.01  & {\showPerfChange{51.45}{44.32}} & 44.32 & 46.36 & 58.05 & 12.03 \\
    Cow         & {\showPerfChange{59.49}{48.79}} & 48.79 & {\showPerfChange{84.89}{71.19}} & 71.19 & 84.23 & 86.48 & 43.98 \\
    Diningtable & {\showPerfChange{37.18}{35.17}} & 35.17 & {\showPerfChange{51.83}{50.47}} & 50.47 & 49.19 & 63.28 & 18.13 \\
    Dog         & {\showPerfChange{45.88}{51.09}} & 51.09 & {\showPerfChange{84.30}{83.61}} & 83.61 & 84.93 & 87.96 & 43.98 \\
    Horse       & {\showPerfChange{29.47}{68.04}} & 68.04 & {\showPerfChange{82.28}{80.57}} & 80.57 & 80.91 & 86.87 & 36.71 \\
    Motorbike   & {\showPerfChange{56.45}{53.98}} & 53.98 & {\showPerfChange{80.10}{76.07}} & 76.07 & 72.86 & 81.36 & 51.68 \\
    Person      & {\showPerfChange{46.86}{13.37}} & 13.37 & {\showPerfChange{75.25}{66.16}} & 66.16 & 71.41 & 81.50 & 19.04 \\
    Pottedplant & {\showPerfChange{22.72}{14.09}} & 14.09 & {\showPerfChange{44.67}{37.86}} & 37.86 & 63.17 & 49.03 & 14.97 \\
    Sheep       & {\showPerfChange{37.12}{42.38}} & 42.38 & {\showPerfChange{76.00}{69.22}} & 69.22 & 69.85 & 81.79 & 34.12 \\
    Sofa        & {\showPerfChange{34.07}{9.78}}  & 9.78  & {\showPerfChange{69.85}{62.62}} & 62.62 & 62.78 & 72.91 & 42.74 \\
    Train       & {\showPerfChange{59.01}{5.21}}  & 5.21  & {\showPerfChange{84.95}{82.87}} & 82.87 & 82.00 & 85.45 & 23.04 \\
    TV  & {\showPerfChange{28.50}{0.90}}  & 0.90  & {\showPerfChange{74.83}{71.48}} & 71.48 & 70.82 & 79.94 & 33.36 \\
    \midrule
    \textbf{mAP@0.5}       & {\showPerfChange{44.85}{35.14}} & 35.14 & {\showPerfChange{74.51}{68.14}} & 68.14 & 68.16 & 77.80 & 33.23 \\
    \textbf{AP@[0.5:0.95]} & {\showPerfChange{22.58}{19.52}} & 19.52 & {\showPerfChange{45.32}{41.48}} & 41.48 & 38.46 & 49.76 & {-} \\
    \textbf{AP@0.75}       & {\showPerfChange{19.32}{17.84}} & 17.84 & {\showPerfChange{47.61}{42.14}} & 42.14 & 38.39 & 52.36 & {-} \\
    \bottomrule
    \multicolumn{8}{l}{\footnotesize $^{\dagger}$ Re-annotated ImaginaryNet. \quad $^{\ddagger}$ Original ImaginaryNet with WSOD.}
  \end{tabular}
  }
\vspace{-1.5em}
\end{table*}

\subsection{Main Results}
We compare Cascade HQP-DETR against DETR-based baselines, trained on FluxVOC and evaluated on VOC 2007 test \cite{pascalvoc} (Table~\ref{tab:main_results}). Most baselines are trained for 30 epochs, while our method uses only 12 epochs to demonstrate training efficiency.

Cascade HQP-DETR achieves 61.04\% mAP@0.5, surpassing the strongest ResNet-50-based baseline StableDINO \cite{stableDino} (57.28\%) by +3.76 points despite 60\% fewer training epochs. Remarkably, our method even slightly outperforms DEIMv2-S \cite{deimv2}(60.96\% mAP@0.5), which leverages the significantly more powerful DinoV3 \cite{dinov3} backbone (shown in gray in Table~\ref{tab:main_results}), demonstrating the effectiveness of our architectural innovations. The advantage is consistent across stricter metrics: 40.75\% vs. 38.27\% (+2.48) on mean AP compared to StableDINO, and 43.68\% vs. 40.42\% (+3.26) on AP@0.75.

\begin{table*}[!t]
  \centering
  \caption{Performance on VOC 2007 test set \cite{pascalvoc} (mAP@0.5, \%) with models trained on FluxVOC. \textit{q}: query count, \textit{ep}: training epochs. \textcolor{darkgray}{Gray text} indicates results using the stronger DinoV3 \cite{dinov3} backbone.}
  \label{tab:main_results}

  \begingroup
  \small
  \renewcommand{\arraystretch}{0.75}
  \resizebox{\textwidth}{!}{%
    \begin{tabular}{p{1.3cm} S[table-format=2.2] S[table-format=2.2] S[table-format=2.2] S[table-format=2.2] S[table-format=2.2] S[table-format=2.2] S[table-format=2.2] S[table-format=2.2] S[table-format=2.2] S[table-format=2.2] S[table-format=2.2] S[table-format=2.2]}
      \toprule
      \textbf{Category} &
      \multicolumn{3}{c}{\textbf{Standard DETRs}} &
      \multicolumn{2}{c}{\textbf{Deformable DETRs}} &
      \multicolumn{5}{c}{\textbf{Advanced Baselines}} &
      \multicolumn{2}{c}{\textbf{Ours}} \\
      \cmidrule(lr){2-4} \cmidrule(lr){5-6} \cmidrule(lr){7-11} \cmidrule(lr){12-13}

      & {\makecell[c]{\textbf{DETR}\\\small\textit{100q,}\\\small\textit{50ep}}} &
      {\makecell[c]{\textbf{DAB}\\\textbf{DETR}\\\small\textit{300q,}\\\small\textit{30ep}}} &
      {\makecell[c]{\textbf{DN}\\\textbf{DETR}\\\small\textit{300q,}\\\small\textit{30ep}}} &
      {\makecell[c]{\textbf{DAB}\\\textbf{Def.}\\\small\textit{300q,}\\\small\textit{30ep}}} &
      {\makecell[c]{\textbf{DN}\\\textbf{Def.}\\\small\textit{300q,}\\\small\textit{30ep}}} &
      {\makecell[c]{\textbf{DINO}\\\small\textit{300q,}\\\small\textit{30ep}}} &
      {\makecell[c]{\textbf{Stable}\\\textbf{DINO}\\\small\textit{300q,}\\\small\textit{30ep}}} &
      {\makecell[c]{\textbf{DEIMv2}\\\textbf{Femto}\\\small\textit{300q,}\\\small\textit{30ep}}} &
      {\makecell[c]{\textbf{DEIMv2}\\\textbf{Nano}\\\small\textit{300q,}\\\small\textit{30ep}}} &
      {\textcolor{darkgray}{\makecell[c]{\textbf{DEIMv2-S}\\\textbf{(DinoV3)}\\\small\textit{300q,}\\\small\textit{12ep}}}} &
      {\makecell[c]{\textbf{Cascade}\\\textbf{HQP-DETR}\\\small\textit{$N_p$,}\\\small\textit{12ep}}} &
      {\makecell[c]{\textbf{+SAM}\\\textbf{Refine}\\\small\textit{post}\\\small\textit{proc}}} \\
      \midrule

      Aeroplane & 25.86 & 53.67 & 50.73 & 53.00 & 53.83 & 61.65 & 64.11 & 30.21 & 61.26 & \textcolor{darkgray}{66.07} & 68.89 & \textbf{68.93} \\
      Bicycle & 34.36 & 61.29 & 63.47 & 55.12 & 57.11 & 68.05 & 67.32 & 47.09 & 68.86 & \textcolor{darkgray}{75.40} & 72.90 & \textbf{72.95} \\
      Bird & 16.33 & 44.85 & 39.42 & 44.77 & 49.11 & 57.00 & 56.79 & 12.52 & 38.50 & \textcolor{darkgray}{59.60} & 59.24 & \textbf{59.25} \\
      Boat & 13.93 & 33.83 & 38.11 & 38.10 & 37.78 & 45.65 & \textbf{53.61} & 19.28 & 39.18 & \textcolor{darkgray}{52.07} & 50.49 & 50.48 \\
      Bottle & 10.57 & 31.47 & 28.78 & 33.87 & 33.90 & 44.82 & \textbf{48.15} & 11.25 & 39.90 & \textcolor{darkgray}{48.41} & 48.13 & 48.15 \\
      Bus & 50.35 & 66.44 & 67.07 & 63.03 & 62.31 & 70.86 & 75.17 & 52.23 & 75.45 & \textcolor{darkgray}{74.14} & 78.10 & \textbf{78.16} \\
      Car & 32.60 & 63.13 & 66.11 & 63.50 & 62.56 & 73.93 & 77.53 & 60.52 & \textbf{80.20} & \textcolor{darkgray}{85.13} & 73.70 & 73.65 \\
      Cat & 28.17 & 57.35 & 44.48 & 50.70 & 51.91 & 62.60 & 66.81 & 20.53 & 37.05 & \textcolor{darkgray}{70.45} & 69.66 & \textbf{69.71} \\
      Chair & 16.91 & 28.32 & 27.42 & 27.75 & 29.69 & 31.08 & 32.84 & 14.14 & 34.30 & \textcolor{darkgray}{44.54} & 40.12 & \textbf{40.18} \\
      Cow & 20.20 & 59.49 & 55.26 & 58.19 & 55.79 & 67.15 & \textbf{71.62} & 33.23 & 66.95 & \textcolor{darkgray}{67.90} & 69.83 & 69.85 \\
      Diningtable & 23.81 & 37.18 & 41.19 & 32.84 & 39.24 & 42.05 & 45.06 & 12.62 & 43.59 & \textcolor{darkgray}{38.11} & 53.12 & \textbf{53.13} \\
      Dog & 10.00 & 45.88 & 34.93 & 40.74 & 45.18 & 42.39 & 45.92 & 14.88 & 32.02 & \textcolor{darkgray}{62.26} & 54.61 & \textbf{54.66} \\
      Horse & 13.99 & 29.47 & 33.21 & 35.69 & 37.17 & 48.07 & 50.34 & 15.69 & 38.25 & \textcolor{darkgray}{59.11} & 71.47 & \textbf{71.50} \\
      Motorbike & 30.72 & 56.45 & 60.61 & 54.21 & 58.70 & 64.20 & 66.28 & 43.37 & 65.62 & \textcolor{darkgray}{70.62} & 71.98 & \textbf{72.03} \\
      Person & 19.78 & 46.86 & 47.89 & 56.39 & 57.76 & 65.04 & 67.53 & 44.42 & 70.19 & \textcolor{darkgray}{74.58} & 73.45 & \textbf{73.48} \\
      Pottedplant & 10.49 & 22.72 & 20.28 & 16.01 & 21.33 & 27.10 & 28.29 & 10.14 & 17.29 & \textcolor{darkgray}{36.18} & 30.32 & \textbf{30.33} \\
      Sheep & 18.73 & 37.12 & 53.52 & 51.61 & 50.62 & 65.20 & \textbf{66.33} & 30.43 & 58.70 & \textcolor{darkgray}{65.03} & 60.42 & 60.40 \\
      Sofa & 32.85 & 34.07 & 33.85 & 43.62 & 39.29 & 50.23 & 49.80 & 20.70 & 41.08 & \textcolor{darkgray}{61.47} & 54.69 & \textbf{54.74} \\
      Train & 52.20 & 59.01 & 69.02 & 52.73 & 60.99 & 72.88 & \textbf{75.32} & 48.74 & 65.78 & \textcolor{darkgray}{69.58} & 72.94 & 72.98 \\
      TV & 15.47 & 28.50 & 36.91 & 29.73 & 29.22 & 30.96 & 36.76 & 11.80 & 26.76 & \textcolor{darkgray}{38.45} & \textbf{46.12} & 46.09 \\
      \midrule

      \textbf{mAP@0.5} & 23.87 & 44.85 & 45.61 & 45.08 & 46.68 & 54.55 & 57.28 & 28.96 & 50.04 & \textcolor{darkgray}{60.96} & 61.01 & \textbf{61.04} \\
      \textbf{mean AP} & 9.70 & 22.58 & 21.72 & 29.45 & 30.90 & 36.80 & 38.27 & 16.50 & 33.36 & \textcolor{darkgray}{41.82} & 40.71 & \textbf{40.75} \\
      \textbf{AP@0.75} & 7.02 & 19.32 & 18.23 & 30.21 & 32.31 & 38.81 & 40.42 & 16.32 & 35.82 & \textcolor{darkgray}{45.83} & 43.67 & \textbf{43.68} \\
      \bottomrule
    \end{tabular}%
  }
  \endgroup
\end{table*}

\begin{figure}[htbp]
    \centering 
    \includegraphics[width=0.9\linewidth]{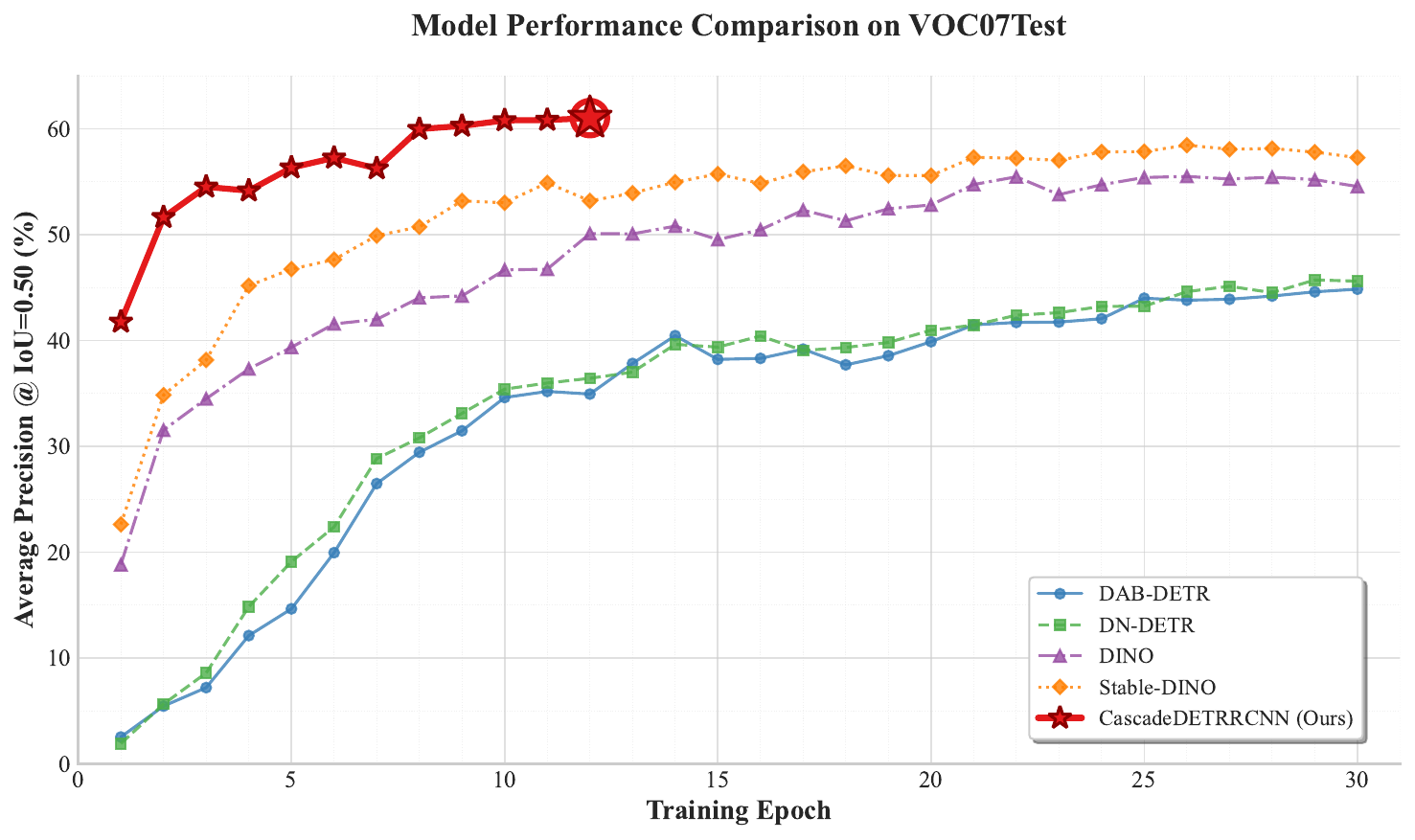}
    \caption{Training convergence on VOC 07 test. Our model converges to higher accuracy in 12 epochs vs. baselines' 30 epochs, demonstrating the effectiveness of HQP initialization.}
    \label{fig:model_comparison_AP50}
\end{figure}

Fig.~\ref{fig:model_comparison_AP50} illustrates the training dynamics. Our model starts at 41.76\% mAP in epoch 1— already competitive with baselines' early-stage performance—and converges to 61.01\% by epoch 12. In contrast, all baselines require 30 epochs to reach lower final accuracies. This fast convergence validates that HQP initialization provides effective image-specific priors, enabling the model to focus on learning transferable representations rather than exploring query distributions from scratch.

\subsubsection{Performance on Real-World Data.}
To assess whether our architectural improvements generalize beyond ISOD, we train all models on VOC 2012 trainval and evaluate on VOC 07 test (Table~\ref{tab:main_results_real_data}).
Cascade HQP-DETR achieves 79.40\% mAP@0.5, 58.28\% mean AP, and 64.03\% AP@0.75, outperforming StableDINO \cite{stableDino} (79.10\%, 58.10\%, 63.91\%) by +0.30, +0.18, and +0.12 points respectively. Notably, our method achieves competitive performance with DEIMv2-S \cite{deimv2} (79.45\% mAP@0.5, 58.94\% mean AP, 64.91\% AP@0.75) despite the latter leveraging the significantly more powerful DinoV3 \cite{dinov3} backbone. While the gains are modest compared to strong baselines, they demonstrate that our method remains effective even when trained on real data with clean annotations and less domain gap—validating its general applicability beyond synthetic-to-real transfer scenarios.

\begin{table*}[!t]
  \centering
  \caption{Performance on VOC 07 test with models trained on real VOC 2012 trainval. \textcolor{darkgray}{Gray text} indicates DEIMv2-S using DinoV3 backbone. Best results in each row are highlighted in bold.}
  \label{tab:main_results_real_data}
  
  \begingroup
  \small
  \renewcommand{\arraystretch}{0.75}
  
  \resizebox{\textwidth}{!}{%
    \begin{tabular}{p{1.3cm} S[table-format=2.2] S[table-format=2.2] S[table-format=2.2] S[table-format=2.2] S[table-format=2.2] S[table-format=2.2] S[table-format=2.2] S[table-format=2.2] S[table-format=2.2] S[table-format=2.2] S[table-format=2.2] S[table-format=2.2]}
      \toprule
      \textbf{Category} & 
      \multicolumn{3}{c}{\textbf{Standard DETRs}} & 
      \multicolumn{2}{c}{\textbf{Deformable DETRs}} & 
      \multicolumn{5}{c}{\textbf{Advanced Baselines}} &
      \multicolumn{2}{c}{\textbf{Ours}} \\
      \cmidrule(lr){2-4} \cmidrule(lr){5-6} \cmidrule(lr){7-11} \cmidrule(lr){12-13}
      
      & {\makecell[c]{\textbf{DETR}\\\small\textit{100q,}\\\small\textit{30ep}}} & 
      {\makecell[c]{\textbf{DAB}\\\textbf{DETR}\\\small\textit{300q,}\\\small\textit{30ep}}} & 
      {\makecell[c]{\textbf{DN}\\\textbf{DETR}\\\small\textit{300q,}\\\small\textit{30ep}}} & 
      {\makecell[c]{\textbf{DAB}\\\textbf{Def.}\\\small\textit{300q,}\\\small\textit{30ep}}} & 
      {\makecell[c]{\textbf{DN}\\\textbf{Def.}\\\small\textit{300q,}\\\small\textit{30ep}}} & 
      {\makecell[c]{\textbf{DINO}\\\small\textit{300q,}\\\small\textit{30ep}}} & 
      {\makecell[c]{\textbf{Stable}\\\textbf{DINO}\\\small\textit{300q,}\\\small\textit{30ep}}} & 
      {\makecell[c]{\textbf{DEIMv2}\\\textbf{Femto}\\\small\textit{300q,}\\\small\textit{30ep}}} &
      {\makecell[c]{\textbf{DEIMv2}\\\textbf{Nano}\\\small\textit{300q,}\\\small\textit{30ep}}} &
      {\textcolor{darkgray}{\makecell[c]{\textbf{DEIMv2-S}\\\textbf{(DinoV3)}\\\small\textit{300q,}\\\small\textit{12ep}}}} &
      {\makecell[c]{\textbf{Cascade}\\\textbf{HQP-DETR}\\\small\textit{$N_p$,}\\\small\textit{12ep}}} & 
      {\makecell[c]{\textbf{+SAM}\\\textbf{Refine}\\\small\textit{post}\\\small\textit{proc}}} \\
      \midrule
      
      Aeroplane   & 16.12 & 75.06 & 74.19 & 80.58 & 82.67 & 83.23 & 84.89 & 74.44 & \textbf{87.25} & \textcolor{darkgray}{85.18} & 85.12 & 85.13 \\
      Bicycle     & 17.65 & 77.92 & 75.24 & 77.34 & 78.40 & 79.87 & 82.74 & 70.29 & \textbf{86.26} & \textcolor{darkgray}{85.64} & 82.95 & 82.98 \\
      Bird        & 11.56 & 71.21 & 76.12 & 80.26 & 78.37 & 81.08 & 79.95 & 60.32 & \textbf{82.90} & \textcolor{darkgray}{86.37} & 80.30 & 80.32 \\
      Boat        & 11.10 & 53.46 & 55.56 & 61.15 & 64.89 & 69.08 & 70.61 & 58.13 & \textbf{72.05} & \textcolor{darkgray}{73.22} & 71.03 & 70.97 \\
      Bottle      & 3.27  & 46.02 & 49.67 & 60.76 & 60.37 & 64.16 & 68.16 & 31.38 & 61.63 & \textcolor{darkgray}{58.35} & 68.71 & \textbf{68.73} \\
      Bus         & 14.43 & 74.03 & 75.96 & 78.80 & 79.56 & 80.84 & 86.04 & 75.38 & \textbf{87.60} & \textcolor{darkgray}{86.14} & 86.03 & 86.02 \\
      Car         & 9.73  & 78.02 & 77.00 & 83.15 & 82.97 & 85.03 & 87.33 & 77.51 & \textbf{88.98} & \textcolor{darkgray}{88.64} & 87.22 & 87.23 \\
      Cat         & 18.86 & 86.84 & 87.28 & 86.81 & 88.03 & 88.48 & 89.43 & 79.81 & 86.15 & \textcolor{darkgray}{92.89} & 89.51 & \textbf{89.52} \\
      Chair       & 3.43  & 46.36 & 44.82 & 52.08 & 54.49 & 54.59 & 59.75 & 38.77 & 59.74 & \textcolor{darkgray}{58.64} & 60.50 & \textbf{60.54} \\
      Cow         & 4.30  & 77.59 & 75.21 & 81.50 & 82.75 & 85.18 & 87.83 & 61.71 & \textbf{88.19} & \textcolor{darkgray}{85.80} & 87.71 & 87.70 \\
      Diningtable & 23.10 & 49.19 & 56.26 & 68.46 & 65.02 & 65.16 & 66.72 & 51.70 & \textbf{69.28} & \textcolor{darkgray}{68.01} & 67.35 & 67.36 \\
      Dog         & 8.71  & 84.23 & 84.78 & 86.87 & 86.57 & 86.55 & 87.63 & 71.45 & 84.70 & \textcolor{darkgray}{89.23} & 87.83 & \textbf{87.84} \\
      Horse       & 18.61 & 80.91 & 79.89 & 80.87 & 82.97 & 79.26 & 85.40 & 69.49 & \textbf{86.26} & \textcolor{darkgray}{85.32} & 85.58 & 85.57 \\
      Motorbike   & 19.29 & 72.86 & 75.93 & 76.83 & 79.23 & 82.68 & 83.06 & 67.94 & \textbf{84.29} & \textcolor{darkgray}{84.59} & 83.31 & 83.30 \\
      Person      & 14.30 & 71.41 & 69.99 & 76.98 & 77.89 & 78.80 & 81.84 & 69.47 & \textbf{83.46} & \textcolor{darkgray}{83.30} & 82.01 & 82.09 \\
      Pottedplant & 1.89  & 34.73 & 35.96 & 46.56 & 44.74 & 49.34 & 52.79 & 31.62 & 48.77 & \textcolor{darkgray}{54.46} & 53.69 & \textbf{53.74} \\
      Sheep       & 6.13  & 69.85 & 67.02 & 77.99 & 75.82 & 78.39 & 82.27 & 66.29 & \textbf{84.32} & \textcolor{darkgray}{82.33} & 82.46 & 82.47 \\
      Sofa        & 15.87 & 62.78 & 66.12 & 64.59 & 70.03 & 68.92 & 75.18 & 62.81 & 73.72 & \textcolor{darkgray}{75.96} & 75.47 & \textbf{75.48} \\
      Train       & 15.19 & 80.00 & 83.04 & 85.88 & 85.33 & 84.51 & 87.53 & 78.79 & 85.06 & \textcolor{darkgray}{88.01} & 87.59 & \textbf{87.60} \\
      TV  & 4.09  & 70.82 & 70.96 & 78.25 & 75.89 & 80.69 & 82.91 & 64.48 & 77.45 & \textcolor{darkgray}{78.87} & \textbf{83.22} & 83.20 \\
      \midrule
      
      \textbf{mAP@0.5}      & 11.88 & 68.16 & 69.05 & 74.29 & 74.80 & 76.29 & 79.10 & 63.09 & 78.90 & \textcolor{darkgray}{79.45} & 79.40 & \textbf{79.41} \\
      \textbf{mean AP}      & 5.77  & 38.46 & 43.11 & 53.61 & 55.00 & 55.97 & 58.10 & 42.05 & 57.41 & \textcolor{darkgray}{58.94} & 58.28 & \textbf{58.29} \\
      \textbf{AP@0.75}      & 5.00  & 38.39 & 44.73 & 57.69 & 59.09 & 60.46 & 63.91 & 45.24 & 63.04 & \textcolor{darkgray}{64.91} & \textbf{64.03} & \textbf{64.03} \\
      \bottomrule
    \end{tabular}%
  }
  \endgroup
\end{table*}

\subsubsection{Qualitative Results.}
\begin{figure*}[!t]
    \includegraphics[width=\textwidth]{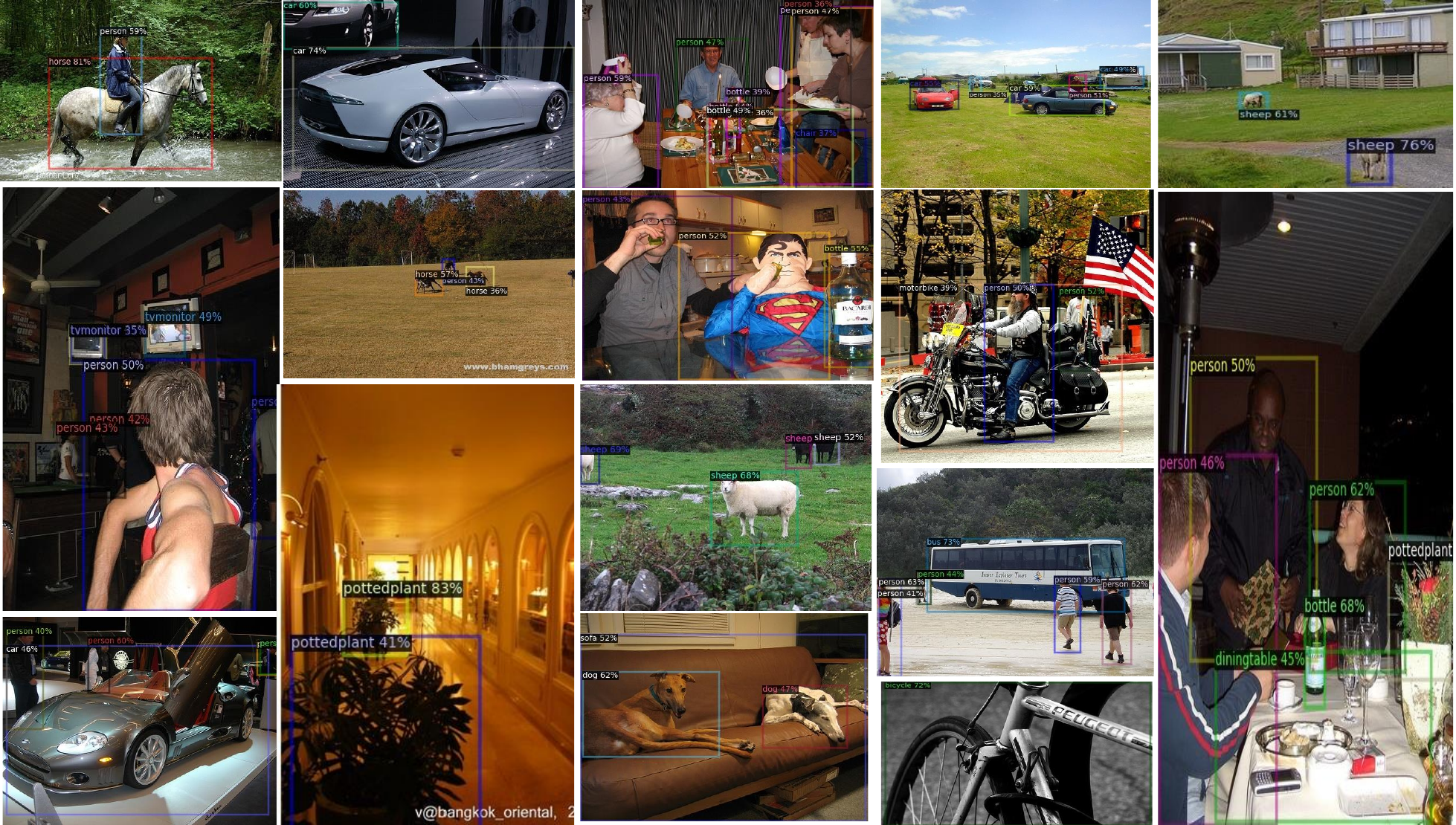}
    \caption{Qualitative detection results of our Cascade HQP-DETR on challenging real images from the PASCAL VOC 07 test set. Trained exclusively on the synthetic \textbf{FluxVOC} dataset, our model demonstrates robust performance on scenes with occlusion, small objects, dense arrangements, and multi-scale variations.}
    \label{fig:result}
\end{figure*}

To visually substantiate our quantitative findings, Fig.~\ref{fig:result} presents detection results on challenging real-world images from the PASCAL VOC07 \cite{pascalvoc} test set. The model was trained exclusively on FluxVOC. The visualizations demonstrate robust sim-to-real generalization across diverse challenging scenarios: heavy \textbf{occlusion} (rider partially obscured by motorcycle), \textbf{small distant objects} (sheep in the field), \textbf{dense arrangements} (dining table scene), and significant \textbf{multi-scale variations}. This strong cross-domain performance demonstrates both the photorealism and diversity of our FluxVOC dataset and our model's generalization.

\subsection{Ablation Study}
We conduct ablation studies on FluxVOC, evaluating on VOC 07 test, to quantify each component's contribution.

\subsubsection{Component Analysis}
Table~\ref{tab:ablation_core_components} presents systematic ablation results starting from DAB-DETR (44.85\% mAP@0.5). We first evaluate positional and content priors separately.
SAM-generated proposals as positional prior yield +11.04 points (55.89\%), highlighting the importance of image-specific geometric guidance. In contrast, RoI-based feature initialization as content prior provides only +1.01 points (45.86\%), suggesting content information is ineffective without spatial localization.Combining them gives a synergistic +13.00 gain (57.85\%). Stable Matching \cite{stableDino} adds +0.87 (58.72\%).Our Cascade DN then outperforms uniform DN by +1.29 points (61.01\% vs 59.72\%). Final SAM refinement adds +0.03 (61.04\%), for a total gain of +16.19.

\begin{table}[!t]
  \centering
  \small
  \caption{Ablation study of core components on the PASCAL VOC07 \cite{pascalvoc} test set. The first row is the DAB-DETR \cite{dabdetr} baseline. Gains are mAP@0.5 improvements relative to this baseline.}
  \label{tab:ablation_core_components}
  
  \begingroup
  \renewcommand{\arraystretch}{0.6}
  
  \resizebox{\textwidth}{!}{%
  \begin{tabular}{cccccc r @{\hskip 1mm}l}
    \toprule
    \textbf{SAM Prop.} & 
    \textbf{RoI Init} & 
    \textbf{Stable Match.} & 
    \textbf{Uniform DN} & 
    \textbf{Cascade DN} & 
    \textbf{SAM Refine} & 
    \multicolumn{2}{c}{\textbf{mAP@0.5 (\%)}} \\
    \midrule
    & & & & & & 44.85 & \\
    \checkmark & & & & & & 55.89 & \tiny{(\textcolor{red}{+11.04})} \\
    & \checkmark & & & & & 45.86 & \tiny{(\textcolor{red}{+1.01})} \\
    \checkmark & \checkmark & & & & & 57.85 & \tiny{(\textcolor{red}{+13.00})} \\
    \checkmark & \checkmark & \checkmark & & & & 58.72 & \tiny{(\textcolor{red}{+13.87})} \\
    \checkmark & \checkmark & \checkmark & \checkmark & & & 59.72 & \tiny{(\textcolor{red}{+14.87})} \\
    \checkmark & \checkmark & \checkmark & & \checkmark & & 61.01 & \tiny{(\textcolor{red}{+16.16})} \\
    \checkmark & \checkmark & \checkmark & & \checkmark & \checkmark & \textbf{61.04} & \tiny{(\textcolor{red}{+16.19})} \\
    \bottomrule
  \end{tabular}
  }
  
  \endgroup 
\end{table}

\begin{table}[!htbp]
  \vspace{-0.3em}
  \centering
  \footnotesize
  \renewcommand{\arraystretch}{0.75}
    \caption{Proposal method comparison on VOC 07 test. Recall at IoU thresholds 0.3/0.5/0.7 (R@X), average proposals per image (Avg. Props), and detection mAP@0.5. All methods use Cascade HQP-DETR.
    }
  \label{tab:proposal_method_analysis}
  \vspace{-0.7em}
  
  \begin{tabular}{
    l
    S[table-format=1.2, round-precision=2]
    S[table-format=1.2, round-precision=2]
    S[table-format=1.2, round-precision=2]
    S[table-format=4.1, round-precision=1]
    S[table-format=2.2]
  }
    \toprule
    \textbf{Method} & 
    {\textbf{R@0.3}} & 
    {\textbf{R@0.5}} & 
    {\textbf{R@0.7}} & 
    {\makecell[c]{\textbf{Avg Props}}} & 
    {\makecell[c]{\textbf{mAP@0.5 (\%)}}} \\
    \midrule
    SAM(32)         & 0.94 & 0.86 & 0.72 & 83.5   & 56.60 \\
    \textbf{SAM(64)} & \textbf{0.99} & \textbf{0.95} & {\textbf{0.79}} & \textbf{181.0} & \textbf{61.01} \\
    SAM(128)        & 0.99 & 0.95 & 0.80 & 219.8  & 59.26 \\
    \midrule
    RPN \cite{fasterrcnn}             & 0.99 & 0.98 & 0.91 & 999.6  & 53.16 \\
    EB \cite{edgebox}              & 0.96 & 0.90 & 0.74 & 1897.0 & 49.09 \\
    SS \cite{ss}              & 0.99 & 0.91 & 0.68 & 1537.4 & 48.57 \\
    \bottomrule
  \end{tabular}%
  \vspace{-0.7em}
\end{table}
\subsubsection{Analysis of Proposal Generation Methods}
We first compare SAM sampling densities to find the optimal setting, then benchmark it against classic methods (RPN \cite{fasterrcnn}, EdgeBoxes \cite{edgebox}, SS \cite{ss}). Table~\ref{tab:proposal_method_analysis} reports recall, average proposals, and final mAP.

\textbf{SAM Configuration.}Among three sampling densities, SAM(64) achieves the best performance at 61.01\% mAP@0.5 with 0.95 recall at IoU 0.5 and 181 proposals per image. SAM(32) suffers from insufficient coverage (recall 0.86), degrading to 56.60\% mAP. SAM(128) provides no recall improvement over SAM(64) but drops to 59.26\%— additional proposals (220 vs 181) introduce redundancy hindering training.

\begin{table}[!htbp]
  \centering
  \footnotesize
  \caption{Cascade denoising hyperparameters ($\theta_1$, $\tau$) on VOC 07 test. Optimal values in bold.}
  \label{tab:hyperparameter_analysis}
  
  \begingroup
  \renewcommand{\arraystretch}{0.75}
  \footnotesize
  
  \begin{tabular}{l c c c c c c c}
    \toprule
    \multicolumn{8}{c}{\textbf{(a) Initial Threshold $\theta_1$}} \\
    \midrule
    \textbf{Threshold $\theta_1$} & 0.1 & 0.2 & \textbf{0.3} & 0.5 & 0.6 & 0.7 & 0.8 \\
    \textbf{mAP@0.5 (\%)} & 60.25 & 60.89 & \textbf{61.01} & 60.90 & 60.75 & 59.98 & 59.95 \\
    \midrule
    \midrule
    \multicolumn{8}{c}{\textbf{(b) Temperature $\tau$}} \\
    \midrule
    \textbf{Temperature $\tau$} & \textbf{0.1} & 0.2 & 0.3 & 0.4 & 0.5 & 0.6 & 0.7 \\
    \textbf{mAP@0.5 (\%)} & \textbf{61.01} & 60.96 & 60.88 & 60.78 & 60.70 & 60.67 & 60.59 \\
    \bottomrule
  \end{tabular}
  \endgroup
\end{table}

\textbf{SAM vs. Classic Methods.} Classic methods generate significantly more proposals: RPN ~1000, EB ~1897, SS ~1537, vs. SAM ~181. Despite higher recall (e.g., RPN: 0.98 at IoU 0.5 vs. SAM's 0.95), their mAP is substantially lower: RPN 53.16\%, EB 49.09\%, SS 48.57\%, all falling below SAM(61.01\%). This gap reveals proposal quality is more critical than quantity or recall. SAM produces compact, object-centric proposals that direct models toward relevant regions, whereas classic methods generate excessive low-quality candidates that introduce training noise.

\subsubsection{Analysis of Cascade Denoising Hyperparameters}
We analyze the sensitivity of two key hyperparameters in our cascade denoising mechanism: initial threshold $\theta_1$ and temperature $\tau$ (Table~\ref{tab:hyperparameter_analysis}).
For $\theta_1$, performance peaks at 0.3 (61.01\%). Lower values (e.g., 0.1: 60.25\%) include too many low-quality queries in early training, causing instability. Higher values (e.g., 0.8: 59.95\%) excessively filter queries, limiting supervision signals. For $\tau$, performance decreases monotonically from 0.1 (61.01\%) to 0.7 (60.59\%). Smaller $\tau$ yields steeper sigmoid curves that sharply distinguish high- and low-quality queries, while larger $\tau$ produces softer weighting with weaker discrimination. These results confirm our choices: $\theta_1 = 0.3$ and $\tau = 0.1$.

\section{Conclusion}
\vspace{-0.8em}
We propose Cascade HQP-DETR, advancing ISOD from weak to full supervision via synthetic data and architectural innovations. We construct FluxVOC and FluxCOCO using LLaMA-3, Flux, and Grounding DINO. Our High-Quality Proposal encoding uses SAM-generated proposals and RoI-pooled features to provide image-specific priors, guiding the model to learn transferable features rather than overfitting to synthetic patterns. Our Cascade Denoising adjusts training pressure via progressively increasing IoU thresholds across decoder layers, encouraging the model to rely on stable visual cues rather than noisy labels. Experiments show our method achieves SOTA when trained on synthetic data and competitive performance on real data.

Despite these advances, limitations remain. First, while HQP encoding improves accuracy, it requires SAM at inference, incurring overhead unsuitable for real-time use. Second, the synthetic-to-real domain gap remains a fundamental challenge, highlighted by the persistent performance gap versus real data training. Third, our data quality is bounded by current Flux and Grounding DINO capabilities. As text-to-image and open-vocabulary detection advance, periodic dataset regeneration may be necessary.

Future research directions emerge from these limitations. For inference efficiency, knowledge distillation could eliminate SAM dependency by training a lightweight proposal network using SAM outputs as teacher signals, preserving quality while reducing cost. Extending to Imaginary Supervised Semantic Segmentation via Grounded-SAM could enable label-free segmentation. Most critically, ISOD's value lies in domains where real data is prohibitively expensive (e.g., industrial defect detection). Validating on such long-tail applications would demonstrate practical impact beyond benchmarks.

\appendix
\section{FluxCOCO Performance on the MS COCO Benchmark}
\label{FluxCOCO Performance on the MS COCO Benchmark}

\begin{figure}[!htbp]
  \centering
  \includegraphics[width=0.85\textwidth]{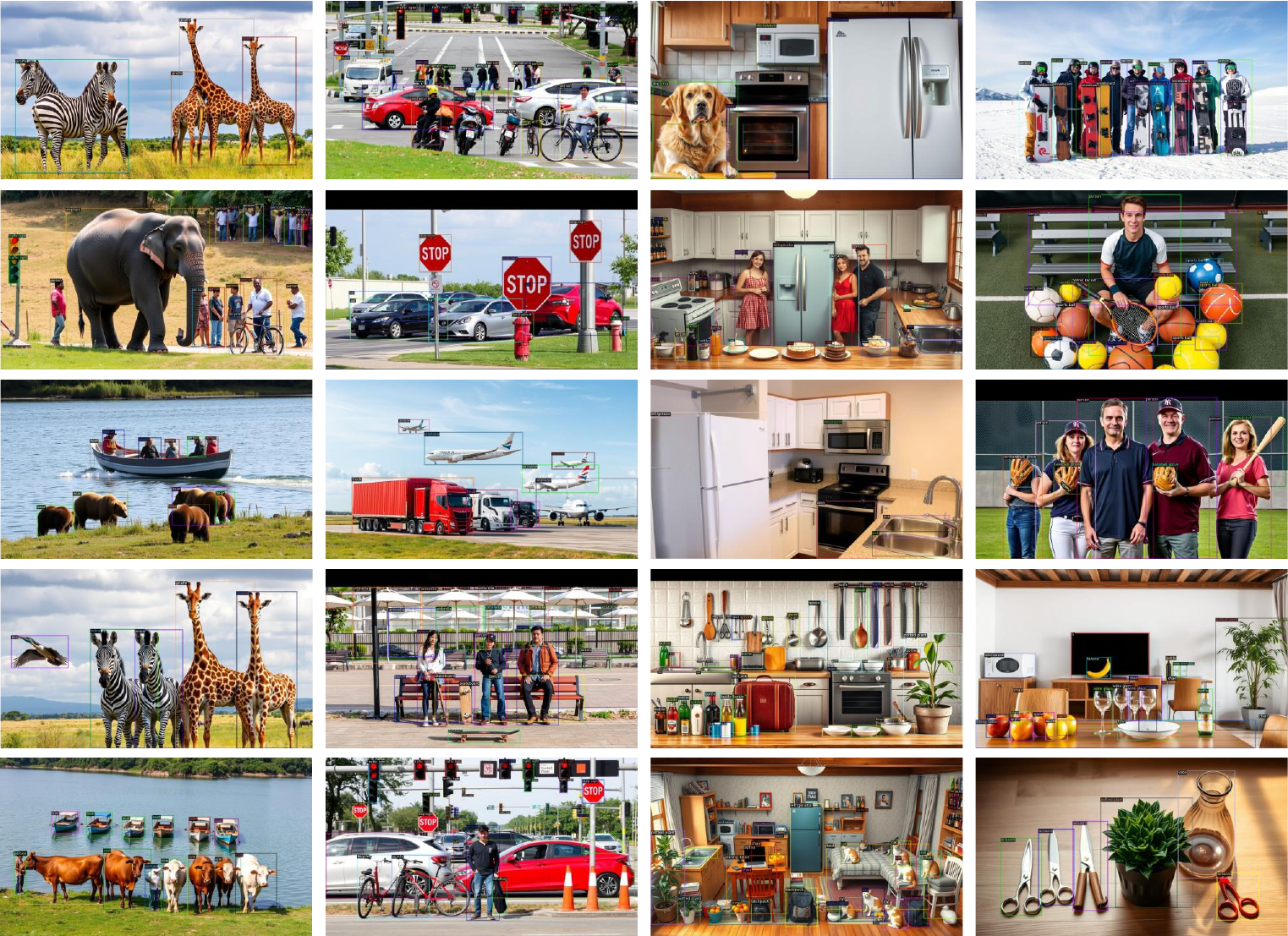}
  \caption{Visual examples of ground-truth annotations from \textbf{FluxCOCO} dataset. The figure highlights the dataset's diversity across various domains. \textbf{Columns (left to right):} (1) Animals (e.g., \textit{zebra, elephant, bear}), (2) Outdoor \& Street Scenes (e.g., \textit{truck, traffic light, fire hydrant}), (3) Indoor Appliances \& Electronics (e.g., \textit{refrigerator, keyboard, sink}), and (4) Sports Equipment \& Accessories (e.g., \textit{snowboard, baseball bat, scissors}). Our model demonstrates robust detection capabilities across these varied and complex scenes.}
  \label{fig:cocodatasets}
  \vspace{-1.5em}
\end{figure}

\begin{table}[!htbp]
  \centering 
  \footnotesize
  \caption{Performance comparison on the MS COCO 2014 validation set. Our model was trained only on the synthetic FluxCOCO dataset. The ImaginaryNet baseline only reported AP$_{50}$.}
  \label{tab:coco_main_results}
  \sisetup{table-format=2.2}
  \begin{tabular}{l S S S S S S}
    \toprule
    \textbf{Model} & {\textbf{AP}} & {\textbf{AP$_{50}$}} & {\textbf{AP$_{75}$}} & {\textbf{AP$_{S}$}} & {\textbf{AP$_{M}$}} & {\textbf{AP$_{L}$}} \\
    \midrule
    ImaginaryNet \cite{imaginarynet} & {-} & 11.40 & {-} & {-} & {-} & {-} \\
    \textbf{Cascade HQP-DETR (Ours)} & \bfseries 20.43 & \bfseries 31.07 & \bfseries 22.03 & \bfseries 8.18 & \bfseries 23.75 & \bfseries 29.55 \\
    \bottomrule
  \end{tabular}
\end{table}

To further validate the scalability and effectiveness of our proposed framework on a more complex and large-scale dataset, we present the results of our \textbf{Cascade HQP-DETR} model trained exclusively on our synthetic \textbf{FluxCOCO} dataset. The model was evaluated on the standard \textbf{MS COCO 2014 validation set} \cite{coco}. To visually illustrate the quality and diversity of our dataset, Figure~\ref{fig:cocodatasets} presents a collage of sample annotated images from FluxCOCO across a range of challenging categories.

Table~\ref{tab:coco_main_results} provides a summary of primary COCO \cite{coco} evaluation metrics, comparing our approach to ImaginaryNet \cite{imaginarynet}. Our method demonstrates substantial improvement, achieving \textbf{20.43\%} overall AP and, critically, \textbf{31.07\% AP}, nearly three times higher than baseline performance. This highlights the significant advantage conferred by our high-quality synthetic data and advanced model architecture.

Table~\ref{tab:coco_per_category} details the per-category AP results for all 80 classes.
\begin{table}[!htbp]
    \centering
    \caption{Per-category AP (\%) on the MS COCO 2014 validation set \cite{coco} for our Cascade HQP-DETR model trained on FluxCOCO.}
    \label{tab:coco_per_category}
    \sisetup{table-format=2.2}
    \footnotesize 
    \resizebox{\textwidth}{!}{%
        \begin{tabular}{lS lS lS lS lS lS}
            \toprule
            \textbf{Category} & {\textbf{AP}} & \textbf{Category} & {\textbf{AP}} & \textbf{Category} & {\textbf{AP}} & \textbf{Category} & {\textbf{AP}} & \textbf{Category} & {\textbf{AP}} & \textbf{Category} & {\textbf{AP}} \\
            \midrule
            person     & 42.57 & horse      & 29.16 & skis         & 4.94  & fork         & 13.15 & bicycle    & 19.64 & sheep        & 29.52 \\
            snowboard  & 11.74 & knife      & 6.34  & car          & 25.54 & cow          & 27.01 & sports ball  & 16.77 & spoon        & 6.79  \\
            motorcycle & 32.36 & elephant   & 36.08 & kite         & 24.16 & bowl         & 18.89 & airplane     & 40.45 & bear         & 33.87 \\
            baseball bat & 14.86 & banana     & 6.00  & bus          & 44.47 & zebra        & 55.83 & baseball glove & 14.49 & apple        & 8.77  \\
            train      & 38.96 & giraffe    & 51.80 & skateboard   & 19.71 & sandwich     & 8.61  & truck        & 19.66 & backpack     & 5.55  \\
            surfboard  & 19.55 & orange     & 11.98 & boat         & 16.93 & umbrella     & 15.17 & tennis racket & 27.74 & broccoli     & 6.72  \\
            traffic light & 18.08 & handbag    & 4.34  & bottle       & 21.36 & carrot       & 3.97  & fire hydrant & 42.64 & tie          & 21.13 \\
            wine glass & 19.12 & hot dog    & 13.72 & stop sign    & 38.76 & suitcase     & 10.47 & cup          & 18.19 & pizza        & 19.83 \\
            parking meter & 10.51 & frisbee    & 35.41 & donut        & 11.95 & cake         & 9.30  & bench        & 13.74 & chair        & 10.34 \\
            couch      & 18.06 & potted plant & 6.55  & bird         & 22.78 & bed          & 24.89 & dining table & 15.44 & toilet       & 29.67 \\
            cat        & 33.17 & tv         & 30.68 & laptop       & 38.95 & mouse        & 28.46 & dog          & 26.36 & remote       & 6.97  \\
            keyboard   & 29.74 & cell phone & 13.59 & microwave    & 30.26 & oven         & 15.42 & toaster      & 8.31  & sink         & 17.20 \\
            refrigerator & 22.61 & book       & 2.44  & clock        & 18.06 & vase         & 19.03 & scissors     & 10.80 & teddy bear   & 25.32 \\
            hair drier & 4.17  & toothbrush & 6.61  &              &       &              &       &              &       &              &       \\
            \bottomrule
        \end{tabular}
    }
\end{table}

\section*{CRediT authorship contribution statement}
\textbf{Zhiyuan Chen}: Conceptualization, Data curation, Formal analysis, Investigation, Methodology, Software, Validation, Visualization, Writing – original draft; \textbf{Yuelin Guo}: Conceptualization, Data curation, Formal analysis, Investigation, Methodology, Project administration, Software, Supervision, Validation, Visualization, Writing – review and editing; \textbf{Zitong Huang}: Validation, Visualization, Writing – review and editing; \textbf{Haoyu He}: Methodology, Supervision, Validation, Visualization, Writing – review and editing; \textbf{Renhao Lu}: Project administration, Resources, Supervision, Writing – review and editing; \textbf{Weizhe Zhang}: Funding acquisition, Project administration, Resources, Supervision, Writing – review and editing

\section*{Data availability}
The code and datasets (FluxVOC, FluxCOCO) are available at \url{https://anonymous.4open.science/r/xeesoxeechen}.

\section*{Declaration of competing interest}
The authors declare that they have no known competing financial interests or personal relationships that could have appeared to influence the work reported in this paper.

\section*{Acknowledgments}
This work was supported in part by the Joint Funds of the National Natural Science Foundation of China (Grant No. \seqsplit{U22A2036}), in part by the National Natural Science Foundation of China (NSFC) / Research Grants Council (RGC) Collaborative Research Scheme  (Grant No. \seqsplit{62461160332} \& \seqsplit{CRS\_HKUST602/24}), in part by the Shenzhen Colleges and Universities Stable Support Program (Grant No. \seqsplit{GXWD20220817124251002}), in part by the Shenzhen Stable Supporting Program (General Project) (Grant No. \seqsplit{GXWD20231130110352002}), in part by the Shenzhen Colleges and Universities Stable Support Program (Grant No. \seqsplit{GXWD20231129102636001}),  in part by the National Natural Science Foundation of China (Grant No. \seqsplit{62402141}), and in part by the Guangdong Basic and Applied Basic Research Foundation (Grant No. \seqsplit{2023A1515110271} \& \seqsplit{2025A1515011785}).

\bibliographystyle{elsarticle-num}
\bibliography{ref}

\end{document}